\documentclass[letterpaper]{article} 
\usepackage{aaai2027}  
\usepackage[hyphens]{url}  
\usepackage{graphicx} 
\usepackage{natbib}  
\usepackage{caption} 
\usepackage{algorithm}
\usepackage{algorithmic}
\usepackage{array}
\usepackage{amsmath}
\usepackage{amssymb}
\usepackage{booktabs}
\usepackage{multirow}
\usepackage{adjustbox}
\usepackage{url}
\usepackage{newfloat}
\usepackage{listings}
\DeclareCaptionStyle{ruled}{labelfont=normalfont,labelsep=colon,strut=off} 
\floatstyle{ruled}
\newfloat{listing}{tb}{lst}{}
\floatname{listing}{Listing}

\title{
An AI4AI Framework for Visual Token Pruning}
\author{
    Zhen Liu\textsuperscript{\rm 1},
    Wenli Huang\textsuperscript{\rm 2}\corresponding,
    Wei Song\textsuperscript{\rm 3},\\
    Yuhan Liu\textsuperscript{\rm 4},
    Zhiqin Yang \textsuperscript{\rm 5}
    Jingwen Fu\textsuperscript{\rm 6}\corresponding
}
\affiliations{
    \textsuperscript{\rm 1} Xi'an Jiaotong University, \textsuperscript{\rm 2} Ningbo University of Technology\\
    \textsuperscript{\rm 3}North China University of Technology, \textsuperscript{\rm 4}MiLM Plus, Xiaomi Inc.\\
    \textsuperscript{\rm 5}Hong Kong University of Science and Technology, \textsuperscript{\rm 6}Zhongguancun Academy\\
    huangwenli@nbut.edu.cn, fujingwen@bza.edu.cn
}

\begin{document}

\maketitle

\begin{abstract}

Visual-token pruning can substantially reduce the inference cost of multimodal large language models (MLLMs), yet existing methods largely rely on fixed, handcrafted heuristics and costly expert trial and error. As pruning objectives, budgets, and model architectures diversify, manually navigating the expanding design space becomes increasingly difficult.
This paper aims to build an \textbf{AI4AI} framework for visual token pruning by solving a natural question: can large language models automatically design effective visual-token reduction algorithms? Although LLMs possess broad algorithmic knowledge and strong reasoning capabilities, translating such general knowledge into effective solutions for a specialized task remains nontrivial. We argue that the key lies in designing an appropriate\textbf{ search-state representation} that connects the internal knowledge of LLMs with the structural requirements and constraints of visual-token pruning.
Based on this insight, we propose \emph{AutoPrune}, a training-free framework for LLM-driven visual-token pruning policy design. At its core, AutoPrune introduces a Token Pruning Domain-Specific Language (TPDSL) comprising 131 reusable atoms for budget control, token scoring, selection constraints, and token reassembly. The \textbf{key property} of TPDSL is to represent each search-state as a residual modification to a strong base policy. This residual formulation narrows the search space and directs the LLM’s attention toward the policy components that are most consequential for performance.
Experiments on 14 multimodal benchmarks and three MLLM backbones demonstrate the effectiveness, efficiency, and transferability of AutoPrune. Even when removing 94.4\% of visual tokens, AutoPrune preserves more than 99\% of full-token performance while reducing FLOPs by $9.9\times$ and prefill latency by $6.4\times$. 
Code is available at \url{https://github.com/AIM-ResearchLab/AutoPrune}.



\end{abstract}


\section{Introduction}

Multimodal large language models (MLLMs) have achieved remarkable progress by integrating powerful language models with visual encoders
\cite{touvron2023llama,bai2023qwen,wang2024qwen2,chen2024internvl}.
Most MLLMs encode an image into hundreds or thousands of visual tokens, substantially increasing inference latency, memory consumption, and deployment cost
\cite{alvar2025divprune}.
Visual-token pruning reduces this overhead by removing redundant or less informative tokens before language-model processing
\cite{chen2024image,xing2024pyramiddrop}. Existing pruning methods largely rely on handcrafted criteria, including attention importance
\cite{xing2024pyramiddrop},
token merging
\cite{shang2025llava},
redundancy estimation
\cite{wen2025stop},
diversity maximization
\cite{alvar2025divprune},
and instruction-conditioned relevance
\cite{zhang2025cdpruner}.
Although effective, designing such policies requires substantial domain expertise and costly trial and error.
As pruning objectives, token budgets, and MLLM architectures become increasingly diverse, manually navigating the resulting design space becomes increasingly difficult.

\begin{figure}[t]
    \centering
    \includegraphics[width=\columnwidth]
    {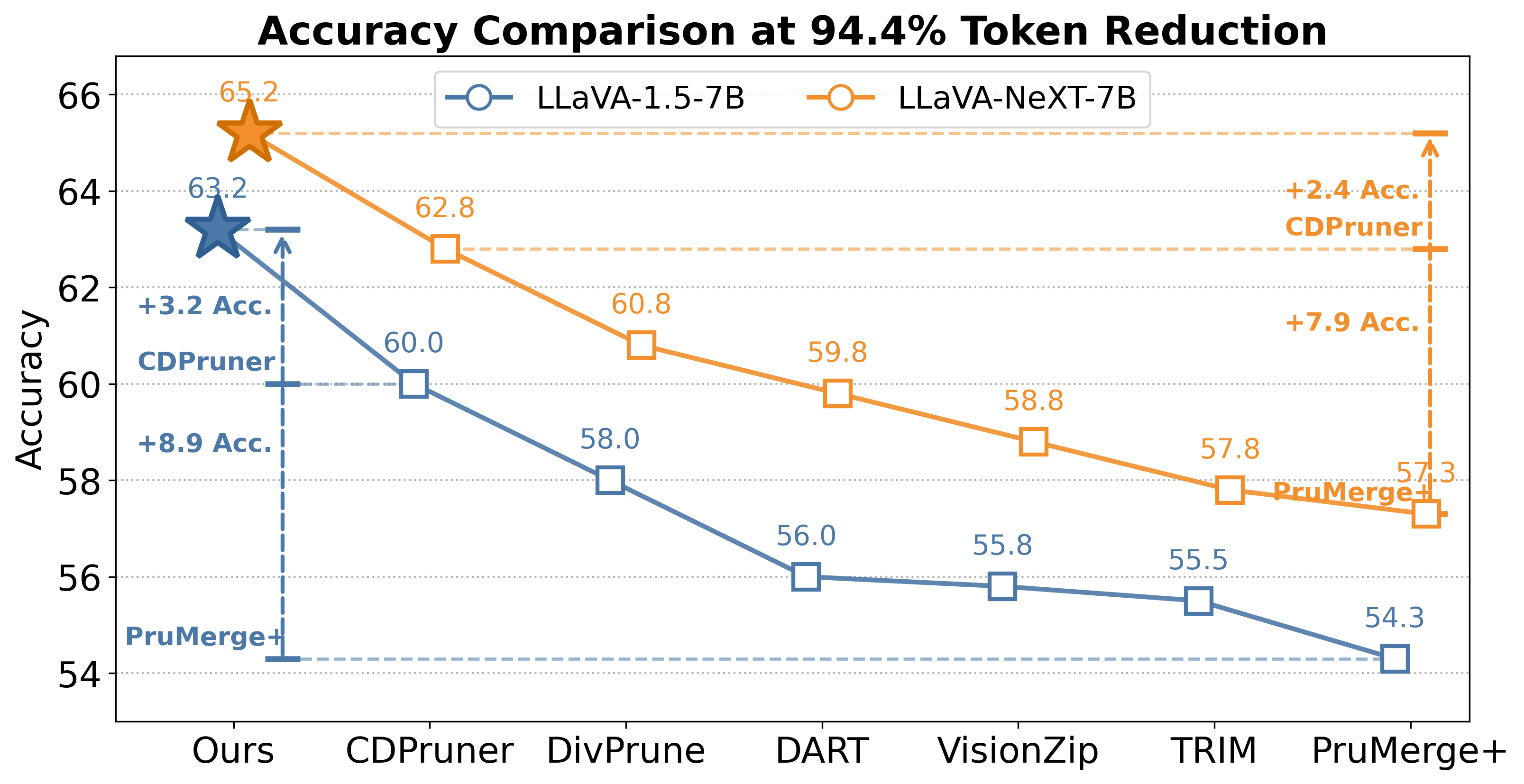}
    \caption{
        Aggregate performance under a 94.4\% visual-token reduction ratio.
        AutoPrune achieves the best performance on both LLaVA-1.5-7B and
        LLaVA-NeXT-7B, outperforming CDPruner~\cite{zhang2025cdpruner}
        by 3.2/2.4 points and PruMerge+~\cite{shang2025llava}
        by 8.9/7.9 points, respectively.
    }
    \label{fig:intro_performance}
\end{figure}

Recent studies have shown that LLMs can generate and improve executable algorithms through iterative proposal and evaluation
\cite{chen2023evoprompting,romera2024funsearch,novikov2025alphaevolve}.
These capabilities offer a promising opportunity to automate visual-token pruning design.
Realizing this opportunity, however, requires more than using an LLM as a code generator.
A central challenge is representational: LLMs possess broad algorithmic knowledge, whereas visual-token pruning is a specialized design problem whose effectiveness depends on the coordination of token scoring, budget allocation, constrained selection, and token reassembly.
Naively representing each candidate as a complete standalone python program forces the LLM to navigate a vast design space, rediscover mechanisms already encoded in existing pruning policies, and satisfy strict pruning constraints simultaneously.
We therefore argue that the key lies in designing an appropriate \emph{search-state representation} that makes the general algorithmic knowledge of LLMs actionable for visual-token pruning.

Based on this insight, we propose \emph{AutoPrune}, a training-free framework for LLM-driven visual-token pruning policy design that requires no fine-tuning of the target MLLM.
AutoPrune introduces a \emph{Token Pruning Domain-Specific Language} (TPDSL) comprising 131 reusable atoms for budget control, token scoring, selection constraints, and token reassembly.
Crucially, TPDSL does not represent a candidate search state as a complete policy designed from scratch.
Instead, each search state combines a strong base policy, a TPDSL-specified residual modification, and the structural constraints required for valid pruning.
This residual formulation preserves reliable prior structure, narrows the effective search space, and directs the LLM toward the policy components that are most consequential for performance. An evaluator-in-the-loop process iteratively generates, validates, and evaluates these residual search states using performance and diagnostic feedback. The selected TPDSL residual specifies how candidate tokens are scored and exchanged with the base-policy selections under a bounded quota, thereby preserving dominant selections while revising only uncertain decisions. By decoupling this residual structure from runtime instantiation, AutoPrune 
supports cross-budget and cross-backbone transfer of the discovered policy.

Experiments on 14 multimodal benchmarks and three MLLM backbones demonstrate the effectiveness, efficiency, and transferability of AutoPrune.
As shown in Figure~\ref{fig:intro_performance}, under a 94.4\% visual-token reduction ratio, AutoPrune preserves more than 99\% of full-token performance on both LLaVA-1.5-7B and LLaVA-NeXT-7B.
In the 320-token efficiency setting on LLaVA-NeXT-7B, AutoPrune reduces FLOPs by $9.9\times$ and prefill latency by $6.4\times$.
Further experiments demonstrate effective transfer across token budgets and MLLM backbones, stable performance across different LLM proposers, and consistent improvements over multiple human-designed base policies.

Our contributions are summarized as follows:

\begin{itemize}
    \item We formulate LLM-driven visual-token pruning as
    residual search around a base policy, using a structured
    search-state representation to make general LLM reasoning
    applicable to a specialized and constraint-intensive
    pruning problem.

    \item We introduce \emph{AutoPrune} and TPDSL to
    instantiate this formulation through 131 reusable pruning
    atoms, evaluator-guided search, quota-constrained residual
    execution, and search-free transfer across token budgets
    and MLLM backbones.

    \item Extensive experiments on 14 multimodal benchmarks
    and three MLLM backbones demonstrate strong pruning
    performance, substantial inference acceleration, effective
    cross-budget and cross-backbone transfer, and robust
    generalization across LLM proposers and base policies.
\end{itemize}

\section{Related Work}

\subsection{Visual-Token Pruning}

Visual-token pruning reduces MLLM inference cost by removing or compressing 
redundant visual tokens. 
Early efficient Vision Transformers explored learned token selection, 
latency-aware pruning, and token squeezing
\cite{kim2022learned,kong2022spvit,wei2023joint,liu2024revisiting}. 
Recent MLLM-oriented methods can be broadly grouped into three categories. 
\emph{Importance-based methods}, such as FastV~\cite{chen2024image}, 
PyramidDrop~\cite{xing2024pyramiddrop}, and SparseVLM~\cite{zhang2024sparsevlm}, 
retain tokens according to attention, saliency, or instruction-conditioned 
relevance~\cite{ye2024fit,sun2025lvpruning}. 
\emph{Redundancy-reduction methods}, including PruMerge~\cite{shang2025llava}, 
TRIM~\cite{song2025less}, and VisionZip~\cite{yang2025visionzip}, merge or 
compress visually similar tokens to reduce duplication
\cite{huang2025prunevid}. 
\emph{Diversity-aware methods}, such as DART~\cite{wen2025stop}, 
DivPrune~\cite{alvar2025divprune}, and CDPruner~\cite{zhang2025cdpruner}, 
seek broader information coverage through redundancy penalties, diversity 
constraints, or instruction-conditioned selection.

Although effective, these policies are largely handcrafted and require 
task-specific tuning. 
AutoPrune instead automates pruning-policy design through structured residual 
search around a strong base policy.

\subsection{LLM-Driven Algorithm Design}

LLMs have increasingly been used to generate and optimize algorithms through 
iterative proposal and evaluation~\cite{liu2026instruction,liu2026elva,liu2026graphir}. 
In neural architecture search, EvoPrompting~\cite{chen2023evoprompting} and 
LLMatic~\cite{nasir2023llmatic} employ LLMs as architecture generators or 
mutation operators. 
FunSearch~\cite{romera2024funsearch} combines program generation with an 
automatic evaluator for mathematical discovery, while 
AlphaEvolve~\cite{novikov2025alphaevolve} extends evaluator-guided evolution 
to broader algorithmic and engineering problems. 
Related studies have also applied LLMs to heuristic generation
\cite{liu2024eoh,van2024llamea} and automated machine-learning pipelines
\cite{trirat2024automl,chi2024sela}.

Direct application to visual-token pruning remains difficult because candidate 
policies must satisfy strict budget, tensor, index, and numerical constraints. 
AutoPrune addresses this challenge with Token Pruning 
Domain-Specific Language (TPDSL), which represents each search 
state as a constrained residual modification to a base policy. 
Together with evaluator-in-the-loop validation and selection, this formulation 
enables executable, budget-compliant, and transferable pruning-policy search.

\begin{figure}[t]
	\centering
	\includegraphics[width=1.01\columnwidth]{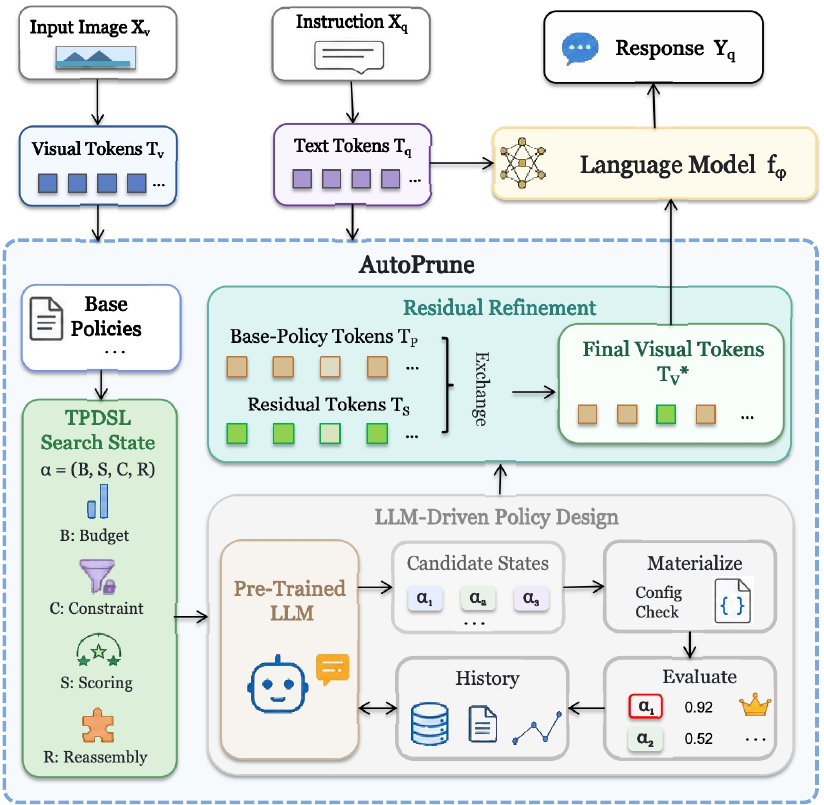}
	\caption{
        Overview of \textbf{AutoPrune}. 
        TPDSL represents each candidate as a structured residual search state
		$\alpha=(B,S,C,R)$ over a strong base policy, enabling executable and
		budget-compliant policy design.
		A pre-trained LLM iteratively proposes, validates, and evaluates candidate
		states using search-history feedback.
		The selected state performs constrained residual refinement, preserving dominant selections while revising only uncertain
decisions.
	}
	
	\label{fig:method_pipeline}
\end{figure}

\section{Method}
\textbf{AutoPrune} formulates visual-token pruning as a constrained 
LLM-driven autodesign problem. 
Instead of generating complete pruning programs from scratch, AutoPrune 
represents each candidate as a TPDSL-structured residual search state defined 
relative to a strong base policy. 
This representation explicitly encodes pruning operations and execution 
constraints, thereby narrowing the search space while preserving the reliable 
structure of the base policy. 
The selected search state is subsequently instantiated through constrained 
residual refinement, which revises a number of uncertain 
base-policy selections. 
Figure~\ref{fig:method_pipeline} illustrates the overall framework.

\subsection{Overall Framework}

Given an input image $X_v$ and a language instruction $X_q$, the vision 
encoder and tokenizer produce visual and text token sequences:
\begin{equation}
\mathcal{T}_v
=
\{T_{v1},T_{v2},\ldots,T_{vN}\},
\mathcal{T}_q
=
\{T_{q1},T_{q2},\ldots,T_{qM}\},
\end{equation}
where $N$ and $M$ denote the numbers of visual and text tokens, respectively. 
Given a target visual-token budget $K\ll N$, a pruning policy $\pi$ selects a retained 
visual-token subset:
\begin{equation}
\mathcal{T}_v^\ast
=
\pi(\mathcal{T}_v,X_q;K),
\qquad
|\mathcal{T}_v^\ast|=K.
\label{eq:pruning_objective}
\end{equation}
The retained visual tokens are concatenated with the text tokens and processed 
by the language model $f_\phi$:
\begin{equation}
Y_q
=
f_\phi
\left(
[\mathcal{T}_q;\mathcal{T}_v^\ast]
\right).
\label{eq:mllm_inference}
\end{equation}

AutoPrune comprises three components. 
First, the \emph{TPDSL search-state representation} expresses each candidate 
as a structured residual modification to a strong base policy, covering budget 
control, token scoring, selection constraints, and token reassembly. 
Second, \emph{LLM-driven policy design} searches these structured states 
through iterative proposal, validation, and task evaluation. 
Third, \emph{prior-guided residual refinement} applies the selected state by 
preserving most base-policy tokens and replacing only a bounded number of 
uncertain selections.

\subsection{TPDSL Search-State Representation.}

The central design of AutoPrune is the \emph{Token Pruning Domain-Specific 
Language} (TPDSL). 
Rather than representing a candidate as an independent pruning program, TPDSL 
defines it as a residual modification to a base policy $\pi_P$:
\begin{equation}
\mathcal{S}_{\alpha}
=
(\pi_P,\alpha),
\qquad
\alpha=(B,S,C,R),
\label{eq:search_state}
\end{equation}
where $B$, $S$, $C$, and $R$ denote \emph{budget control}, 
\emph{token scoring}, \emph{selection constraints}, and 
\emph{token reassembly}, respectively. 
The base policy provides a reliable initial selection, while $\alpha$ specifies 
how that selection is evaluated, revised, and reassembled.

The four TPDSL components serve complementary functions. 
$B$ defines the target token budget and its runtime instantiation rule. 
$S$ specifies token-quality signals and their fusion, including instruction 
relevance, attention-proxy saliency, spatial centrality, redundancy density, 
and local contrast. 
$C$ defines admissible selection behavior, including reference anchoring 
(i.e., base-policy preservation), residual exchange quotas, candidate-pool 
construction, and diversity-aware selection. 
$R$ restores the retained tokens to the MLLM input and optionally incorporates 
residual information from discarded tokens.

TPDSL contains 131 instantiated atoms organized into nine functional groups:
\begin{equation}
\mathcal{L}
=
\bigcup_{g=1}^{9}
\mathcal{L}^{(g)},
\qquad
|\mathcal{L}|=131,
\label{eq:tpdsl_library}
\end{equation}
covering token source and base policy, budget control, scoring, similarity 
kernels, score fusion, candidate-pool construction, selection constraints, 
reassembly and token processing, and safety checks. 
Each atom has predefined semantics and a standardized runtime interface. 
These typed interfaces constrain the LLM to compose pruning operations that are
compatible with budget, shape, and index requirements.
The complete atom organization is provided in the supplementary material.

A TPDSL specification is converted into an executable pruning policy through
\begin{equation}
\pi_{\alpha}^{(K)}
=
\mathrm{Materialize}
\left(
\alpha,
\pi_P,
K
\right),
\label{eq:materialize}
\end{equation}
where materialization resolves default parameters, binds the residual 
specification to the base policy, and instantiates budget- and model-dependent 
fields. 
The resulting policy is validated by
\begin{equation}
d_{\alpha}^{(K)}
=
\mathrm{SafetyCheck}
\left(
\pi_{\alpha}^{(K)},
K
\right),
\label{eq:safety_check}
\end{equation}
where $d_{\alpha}^{(K)}$ records token-budget consistency, index validity, 
numerical stability, output-shape compatibility, and deterministic execution. 
Invalid candidates are excluded from task evaluation.
TPDSL therefore replaces unconstrained code generation with a compositional, 
interpretable, and execution-safe policy space.

\subsection{LLM-Driven Policy Design}
\label{sec:llm_policy_design}

A pre-trained LLM searches over TPDSL-structured residual modifications rather 
than generating arbitrary pruning code. 
At search round $t$, the LLM receives the TPDSL library $\mathcal{L}$, the 
base policy $\pi_P$, and the accumulated search history 
$\mathcal{H}_{t-1}$, and proposes
\begin{equation}
\mathcal{A}_t
=
\mathrm{LLMPropose}
\left(
\mathcal{L},
\pi_P,
\mathcal{H}_{t-1};
K_0
\right),
\label{eq:llm_propose}
\end{equation}
where $\mathcal{A}_t=\{\alpha_{t,i}\}_{i=1}^{m_t}$ contains $m_t$ candidate 
search states and $K_0$ denotes the source search budget.

Each candidate $\alpha_{t,i}$ is materialized and validated according to 
Eqs.~\eqref{eq:materialize} and~\eqref{eq:safety_check}. 
A valid candidate is then evaluated by
\begin{equation}
e_{t,i}
=
\mathcal{E}
\left(
\pi_{\alpha_{t,i}}^{(K_0)}
\right),
\label{eq:policy_evaluation}
\end{equation}
where $\mathcal{E}$ denotes the task evaluator. 
All valid candidates are collected into a global policy pool:
\begin{equation}
\mathcal{V}
=
\bigcup_{t=1}^{T}
\left\{
\left(
\alpha_{t,i},
\pi_{\alpha_{t,i}}^{(K_0)},
e_{t,i}
\right)
\;\middle|\;
\mathrm{Valid}(d_{t,i})=1
\right\}.
\label{eq:valid_policy_pool}
\end{equation}

After each round, the candidate structures, validation diagnostics, and 
evaluation scores are incorporated into the search history:
\begin{equation}
\mathcal{H}_t
=
\mathrm{Summarize}
\left(
\mathcal{H}_{t-1},
\mathcal{O}_t
\right),
\label{eq:history_update}
\end{equation}
where $\mathcal{O}_t$ denotes the observations collected at round $t$. 
This feedback promotes effective TPDSL compositions while discouraging 
invalid or low-performing designs.

After $T$ rounds, AutoPrune selects the highest-scoring valid candidate:
\begin{equation}
\alpha^\ast
=
\underset{\alpha_{t,i}\in\mathcal{V}}
{\arg\max}
\;
e_{t,i}.
\label{eq:best_policy}
\end{equation}
Here, $\alpha^\ast$ denotes the selected TPDSL residual policy structure rather 
than an unconstrained pruning program; its search procedure and materialized 
atom composition are provided in the supplementary material.

\subsection{TPDSL-Guided Residual Refinement}
\label{sec:residual_refinement}

The selected state
$\alpha^\ast=(B^\ast,S^\ast,C^\ast,R^\ast)$ is instantiated as a bounded 
residual refinement of the base policy.
The budget component $B^\ast$ first resolves the requested target budget; for 
notational simplicity, the instantiated value is denoted by $K$.

Given visual tokens $\mathcal{T}_v$, instruction $X_q$, and target budget $K$, 
the base policy first produces an initial token subset:
\begin{equation}
\mathcal{T}_P
=
\pi_P
\left(
\mathcal{T}_v,
X_q;
K
\right),
\qquad
|\mathcal{T}_P|=K,
\label{eq:base_policy_subset}
\end{equation}
where $\mathcal{T}_P$ denotes the base-policy selections.

The scoring component $S^\ast$ evaluates each visual token using complementary 
quality signals:
\begin{equation}
\mathbf{s}_i
=
\left[
s_i^{\mathrm{rel}},
s_i^{\mathrm{attn}},
s_i^{\mathrm{spa}},
s_i^{\mathrm{red}},
s_i^{\mathrm{con}}
\right],
\label{eq:token_signals}
\end{equation}
corresponding to instruction relevance, attention-proxy saliency, spatial 
centrality, redundancy density, and local contrast, respectively. 
The normalized signals are fused into a token-quality score:
\begin{equation}
g_i
=
\prod_m
\left(
\bar{s}_i^{(m)}
\right)^{w_m},
\label{eq:score_fusion}
\end{equation}
where the fusion weights $w_m$ are specified by $\alpha^\ast$.

Based on these scores, the constraint component $C^\ast$ constructs a 
candidate pool $\mathcal{T}_S$ and performs bounded residual exchange:
\begin{equation}
\left(
\mathcal{T}_D,
\mathcal{T}_A
\right)
=
\mathrm{RE}
\left(
\mathcal{T}_P,
\mathcal{T}_S,
\mathbf{g};
q_e,
r_{\min}
\right),
\label{eq:searched_residual_exchange}
\end{equation}
where $\mathcal{T}_D\subseteq\mathcal{T}_P$ contains low-confidence 
base-policy tokens, while
$\mathcal{T}_A\subseteq\mathcal{T}_S\setminus\mathcal{T}_P$ contains 
high-scoring candidate tokens. 
The exchange quota $q_e$ limits the number of replacements, and $r_{\min}$ 
specifies the minimum number of base-policy tokens to preserve.

The refined token subset is
\begin{equation}
\widetilde{\mathcal{T}}_v
=
\left(
\mathcal{T}_P\setminus\mathcal{T}_D
\right)
\cup
\mathcal{T}_A,
\label{eq:refined_tokens}
\end{equation}
subject to
\begin{equation}
|\mathcal{T}_D|
=
|\mathcal{T}_A|
\leq q_e,
|\mathcal{T}_P\setminus\mathcal{T}_D|
\geq r_{\min},
|\widetilde{\mathcal{T}}_v|
=
K.
\label{eq:residual_exchange}
\end{equation}

Finally, the reassembly component $R^\ast$ restores token ordering and, when 
specified, incorporates residual information from discarded tokens:
\begin{equation}
\mathcal{T}_v^\ast
=
\mathrm{Reassemble}
\left(
\widetilde{\mathcal{T}}_v,
\mathcal{T}_v;
R^\ast
\right),
\qquad
|\mathcal{T}_v^\ast|=K.
\label{eq:token_reassembly}
\end{equation}
This bounded refinement preserves the dominant structure of the base policy 
while allowing the selected TPDSL state to correct a small number of uncertain 
token selections.

\begin{algorithm}[t]
	\caption{TPDSL-Guided Residual Refinement}
	\label{alg:tpdsl_residual}
	\begin{algorithmic}[1]

		\REQUIRE Visual tokens $\mathcal{T}_v$, instruction $X_q$, 
		base policy $\pi_P$, selected TPDSL state
		$\alpha^\ast=(B^\ast,S^\ast,C^\ast,R^\ast)$, target budget $K$

		\ENSURE Refined visual tokens $\mathcal{T}_v^\ast$

		\STATE $\widehat{K}
		\leftarrow
		\mathrm{InstantiateBudget}(B^\ast,K)$

		\STATE $\pi_{\alpha^\ast}^{(\widehat{K})}
		\leftarrow
		\mathrm{Materialize}
		(\alpha^\ast,\pi_P,\widehat{K})$

		\STATE $d
		\leftarrow
		\mathrm{SafetyCheck}
		(\pi_{\alpha^\ast}^{(\widehat{K})},\widehat{K})$

		\IF{$\mathrm{Valid}(d)=0$}
			\STATE \textbf{return}
			$\pi_P(\mathcal{T}_v,X_q;\widehat{K})$
		\ENDIF

		\STATE $\mathcal{T}_P
		\leftarrow
		\pi_P(\mathcal{T}_v,X_q;\widehat{K})$

		\STATE $\mathbf{g}
		\leftarrow
		\mathrm{ScoreTokens}
		(\mathcal{T}_v,X_q;S^\ast)$

		\STATE $\mathcal{T}_S
		\leftarrow
		\mathrm{BuildCandidatePool}
		(\mathcal{T}_v,\mathcal{T}_P,\mathbf{g};C^\ast)$

		\STATE $(q_e,r_{\min})
		\leftarrow
		\mathrm{ResolveConstraints}
		(C^\ast,\widehat{K})$

		\STATE $(\mathcal{T}_D,\mathcal{T}_A)
		\leftarrow
		\mathrm{RE}
		(\mathcal{T}_P,\mathcal{T}_S,\mathbf{g};
		q_e,r_{\min})$

		\STATE $\widetilde{\mathcal{T}}_v
		\leftarrow
		(\mathcal{T}_P\setminus\mathcal{T}_D)
		\cup\mathcal{T}_A$

		\STATE $\mathcal{T}_v^\ast
		\leftarrow
		\mathrm{Reassemble}
		(\widetilde{\mathcal{T}}_v,\mathcal{T}_v;R^\ast)$

		\STATE \textbf{assert}
		$|\mathcal{T}_v^\ast|=\widehat{K}$ and
		$|\mathcal{T}_P\setminus\mathcal{T}_D|\geq r_{\min}$

		\STATE \textbf{return} $\mathcal{T}_v^\ast$

	\end{algorithmic}
\end{algorithm}

\begin{table*}[t]
	\centering
    \caption{
	Performance comparison on LLaVA-1.5-7B and LLaVA-NeXT-7B under a 
	94.4\% visual-token reduction ratio.
}
	\label{tab:reduction94}
	\setlength{\tabcolsep}{2.8pt}
	\renewcommand{\arraystretch}{1.00}
	\begin{adjustbox}{max width=0.95\textwidth}
		\begin{tabular}{
				lc
				cccccc
				cccc
				cc
			}
			\toprule
			\textbf{Method} & \textbf{\# Tokens} & \textbf{VQA$^{\mathrm{V2}}$} & \textbf{GQA} & \textbf{VizWiz} & \textbf{SQA$^{\mathrm{IMG}}$} & \textbf{VQA$^{\mathrm{Text}}$} & \textbf{POPE} & \textbf{MME} & \textbf{MMB$^{\mathrm{EN}}$} & \textbf{MMB$^{\mathrm{CN}}$} & \textbf{MM-Vet} & \textbf{Acc.} & \textbf{Rel.} \\
			\midrule
			\multicolumn{14}{c}{ \textit{LLaVA-1.5-7B: 576 $\rightarrow$ 32 visual tokens ($\downarrow$94.4\%)} } \\
			\midrule
			Full Model & 576 & 78.50 & 61.90 & 50.10 & 69.50 & 58.20 & 85.90 & 1506.50 & 64.70 & 58.10 & 31.30 & 63.40 & 100.0 \\
			\cmidrule(lr){1-14}
			PruMerge+~(2024.05) & 32 & 65.60 & 52.90 & 53.50 & 67.90 & 49.20 & 66.70 & 1236.60 & 55.10 & 45.90 & 24.70 & 54.30 & 86.1 \\
			TRIM~(COLING25) & 32 & 68.60 & 54.50 & 50.70 & 68.10 & 47.60 & 84.90 & 1251.80 & 57.70 & 40.10 & 20.50 & 55.50 & 86.2 \\
			VisionZip~(CVPR25) & 32 & 67.10 & 51.80 & 52.40 & 69.10 & 53.10 & 69.40 & 1251.20 & 57.00 & 50.30 & 25.30 & 55.80 & 88.4 \\
			DART~(2025.02) & 32 & 67.10 & 52.90 & 52.50 & 69.30 & 52.20 & 69.10 & 1273.30 & 58.50 & 50.00 & 25.00 & 56.00 & 88.6 \\
			DivPrune~(CVPR25) & 32 & 71.20 & 54.90 & 53.30 & 68.60 & 52.90 & 81.50 & 1284.90 & 57.60 & 49.10 & 26.30 & 58.00 & 91.3 \\
			CDPruner~(NeurIPS25) & 32 & \underline{73.60} & \underline{57.00} & 53.10 & \underline{69.50} & \underline{53.20} & \textbf{87.90} & \underline{1373.00} & \underline{59.60} & \underline{49.60} & \underline{27.80} & \underline{60.00} & \underline{94.3} \\
			\textbf{Ours} & 32 & \textbf{73.91} & \textbf{57.33} & \textbf{54.27} & \textbf{70.10} & \textbf{53.28} & \underline{87.67} & \textbf{1413.46} & \textbf{70.55} & \textbf{65.90} & \textbf{28.00} & \textbf{63.20} & \textbf{99.7} \\
			\midrule
			\multicolumn{14}{c}{ \textit{LLaVA-NeXT-7B: 2880 $\rightarrow$ 160 visual tokens ($\downarrow$94.4\%)} } \\
			\midrule
			Full Model & 2880 & 81.3 & 62.5 & 55.2 & 67.5 & 60.3 & 86.8 & 1511.8 & 65.8 & 57.3 & 40.0 & 65.2 & 100.0 \\
			\cmidrule(lr){1-14}
			PruMerge+~(2024.05) & 160 & 70.5 & 56.2 & \textbf{57.2} & 66.9 & 50.3 & 71.1 & 1289.6 & 58.0 & 48.9 & 29.3 & 57.3 & 87.7 \\
			TRIM~(COLING25) & 160 & 71.0 & 57.4 & 52.9 & 65.5 & 45.8 & 84.8 & 1275.8 & 61.6 & 45.2 & 29.6 & 57.8 & 87.7 \\
			VisionZip~(CVPR25) & 160 & 71.4 & 55.2 & 55.5 & \textbf{67.9} & 55.0 & 74.9 & 1327.8 & 58.6 & 50.4 & 32.3 & 58.8 & 90.0 \\
			DART~(2025.02) & 160 & 72.5 & 56.8 & \underline{56.7} & \underline{67.8} & 54.9 & 75.3 & 1325.4 & 62.0 & 53.6 & 32.2 & 59.8 & 91.7 \\
			DivPrune~(CVPR25) & 160 & 75.0 & 59.3 & 56.1 & 67.1 & 54.1 & 80.0 & 1356.6 & 62.9 & 53.7 & 32.0 & 60.8 & 92.9 \\
			CDPruner~(NeurIPS25) & 160 & \underline{76.7} & \textbf{60.8} & 55.2 & 67.5 & \underline{55.4} & \textbf{86.8} & \underline{1425.3} & \underline{64.2} & \underline{53.8} & \textbf{36.2} & \underline{62.8} & \underline{96.0} \\
			\textbf{Ours} & 160 & \textbf{76.9} & \underline{60.7} & 56.3 & 67.4 & \textbf{55.6} & \textbf{86.8} & \textbf{1430.4} & \textbf{73.3} & \textbf{67.9} & \underline{35.3} & \textbf{65.2} & \textbf{99.9} \\
			\bottomrule
		\end{tabular}
	\end{adjustbox}
\end{table*}

\begin{figure}[t]
	\centering
	\includegraphics[width=0.98\columnwidth]
	{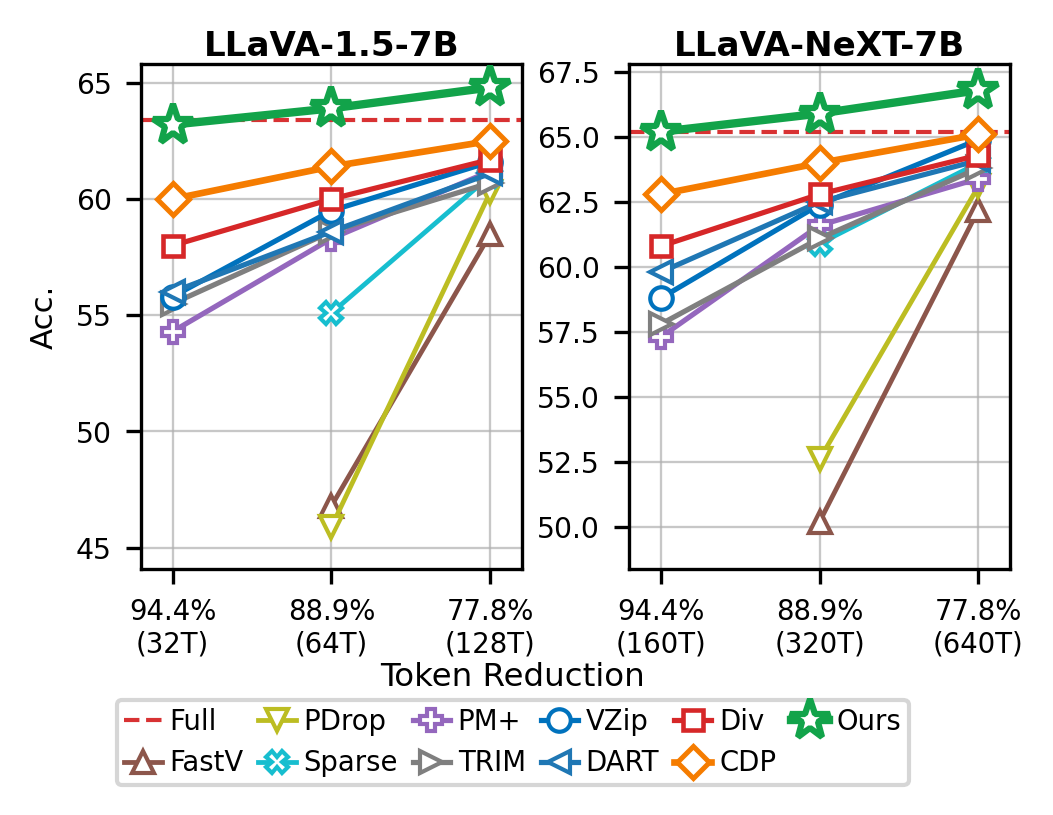}
	\caption{
		Aggregate performance under different visual-token reduction ratios on 
		LLaVA-1.5-7B and LLaVA-NeXT-7B. 
		AutoPrune consistently achieves the best performance, with larger gains 
		under more aggressive pruning.
	}
	\label{fig:performance-image}
\end{figure}

\begin{table}[t]
	\centering
	\scriptsize
	\caption{
		Efficiency performance on LLaVA-NeXT-7B using a single 
		NVIDIA RTX 3090 GPU. All pruning methods retain 320 visual tokens.
	}
	\label{tab:eff}
	
	\setlength{\tabcolsep}{1.0pt}
	\renewcommand{\arraystretch}{1.08}
	
	\begin{tabular}{
			@{}
			>{\raggedright\arraybackslash}p{0.22\columnwidth}
			>{\centering\arraybackslash}p{0.14\columnwidth}
			>{\centering\arraybackslash}p{0.14\columnwidth}
			>{\centering\arraybackslash}p{0.11\columnwidth}
			>{\centering\arraybackslash}p{0.1\columnwidth}
			>{\centering\arraybackslash}p{0.13\columnwidth}
			>{\centering\arraybackslash}p{0.09\columnwidth}
			@{}
		}
		\toprule
			\textbf{Method} & \shortstack{\textbf{FLOPs}\\\textbf{(T)}} & \shortstack{\textbf{Prefill}\\\textbf{(ms/token)}} & \shortstack{\textbf{Dec.}\\\textbf{(ms/token)}} & \shortstack{\textbf{KV Ca.}\\\textbf{(MB)}} & \shortstack{\textbf{GPU Mem.}\\\textbf{(GB)}} & \shortstack{\textbf{Score}\\\textbf{(MME)}} \\
		\midrule
			Full Model & 41.7 & 815 & 43 & 1440.0 & 17.2 & 1511.8 \\
			\shortstack[l]{FastV} & \shortstack{4.4($\times 9.5$)} & \shortstack{186.2($\times 4.3$)} & \shortstack{29} & 160.3 & 15.6 & 1099.0 \\
			\shortstack[l]{PDrop} & \shortstack{4.5($\times 9.3$)} & \shortstack{189.6($\times 4.2$)} & \shortstack{30} & 160.2 & 15.6 & 1171.5 \\
			\shortstack[l]{SparseVLM} & \shortstack{4.5($\times 9.3$)} & \shortstack{244.8($\times 3.3$)} & \shortstack{31} & 161.2 & 18.6 & 1386.1 \\
			\shortstack[l]{VisionZip} & \shortstack{\textbf{4.2}\textbf{($\times 9.9$)}} & \shortstack{131($\times 6.2$)} & \shortstack{\textbf{28}} & \textbf{160.0} & 14.8 & 1397.1 \\
			\shortstack[l]{CDPruner} & \shortstack{\textbf{4.2}\textbf{($\times 9.9$)}} & \shortstack{131($\times 6.2$)} & \shortstack{\textbf{28}} & \textbf{160.0} & \textbf{13.8} & \underline{1453.0} \\
			\textbf{Ours} & \shortstack{\textbf{4.2}\textbf{($\times 9.9$)}} & \shortstack{\textbf{128}\textbf{($\times 6.4$)}} & \shortstack{\textbf{28}} & \textbf{160.0} & \textbf{13.8} & \textbf{1457.9} \\
		\bottomrule
	\end{tabular}
\end{table}

\begin{table*}[t]
	\centering
	\caption{
Cross-backbone transfer to Qwen2.5-VL-7B across visual-token budgets.
	}
	\label{tab:qwen}
	
	\setlength{\tabcolsep}{3.2pt}
	\renewcommand{\arraystretch}{1.00}
	
	\begin{adjustbox}{max width=0.85\textwidth}
		\begin{tabular}{
				lc
				ccccccc
				cc
			}
			\toprule
			\textbf{Method} & \textbf{\# Tokens} & \textbf{TextVQA} & \textbf{ChartQA} & \textbf{AI2D} & \textbf{HallBench} & \textbf{MME} & \textbf{MMB$^{\mathrm{EN}}$} & \textbf{MMB$^{\mathrm{CN}}$} & \textbf{Acc.} & \textbf{Rel.} \\
			\midrule
			Qwen2.5-VL-7B & 1296 & 84.80 & 86.10 & 80.40 & 46.80 & 2304.00 & 82.80 & 83.20 & 78.06 & 100.0 \\
			\midrule
			CDPruner & \multirow{2}{*}{256 ($\downarrow$80.2\%)} & 31.84 & 48.96 & 68.62 & 60.67 & 1887.85 & \textbf{77.75} & 76.29 & 61.65 & 82.4 \\
			\textbf{Ours} & & \textbf{51.69} & \textbf{57.32} & \textbf{69.27} & \textbf{62.46} & \textbf{1974.16} & 76.55 & \textbf{77.15} & \textbf{66.42} & \textbf{88.3} \\
			\midrule
			CDPruner & \multirow{2}{*}{128 ($\downarrow$90.1\%)} & 23.24 & 32.68 & 66.26 & 58.04 & \textbf{1755.08} & 72.85 & 73.28 & 55.58 & 74.9 \\
			\textbf{Ours} & & \textbf{41.42} & \textbf{47.72} & \textbf{66.68} & \textbf{58.99} & 1743.74 & \textbf{73.63} & \textbf{73.88} & \textbf{60.66} & \textbf{81.0} \\
			\midrule
			CDPruner & \multirow{2}{*}{64 ($\downarrow$95.1\%)} & 18.27 & 20.92 & 64.09 & 57.20 & 1596.26 & \textbf{69.07} & 69.07 & 50.80 & 69.1 \\
			\textbf{Ours} & & \textbf{30.40} & \textbf{37.20} & \textbf{65.58} & \textbf{57.83} & \textbf{1657.21} & 68.04 & \textbf{70.45} & \textbf{55.53} & \textbf{74.7} \\
			\midrule
			CDPruner & \multirow{2}{*}{32($\downarrow$97.5\%)} & 14.84 & 16.84 & 62.60 & 57.10 & \textbf{1555.14} & 63.83 & 64.00 & 47.82 & 65.5 \\
			\textbf{Ours} & & \textbf{20.78} & \textbf{28.20} & \textbf{63.47} & \textbf{57.20} & 1492.18 & \textbf{64.69} & \textbf{64.95} & \textbf{50.37} & \textbf{68.5} \\
			\bottomrule
		\end{tabular}
	\end{adjustbox}
\end{table*}

\begin{table}[t]
	\centering
	\small
	\caption{
		Ablation of TPDSL components. 
		$\Delta_{\mathrm{Base}}$ and $\Delta_{\mathrm{Full}}$ are relative to 
		the CDPruner base policy and full-TPDSL variants, respectively.
	}
	\label{tab:ablation_tpdsl_design}
	\setlength{\tabcolsep}{3pt}
	\begin{tabular}{lccc}
		\toprule
		\textbf{Method}
		&
		\textbf{MME}
		&
		\textbf{$\Delta_{\mathrm{Base}}$}
		&
		\textbf{$\Delta_{\mathrm{Full}}$}
		\\
		\midrule
		
		CDPruner base policy
		&
		1373.00
		&
		+0.00
		&
		-40.46
		\\
		
		TPDSL w/o reference anchoring
		&
		1223.29
		&
		-149.71
		&
		-190.17
		\\
		
		TPDSL w/o multi-score fusion
		&
		1391.70
		&
		+18.70
		&
		-21.76
		\\
		
		TPDSL w/o diversity selection
		&
		\underline{1400.80}
		&
		\underline{+27.80}
		&
		\underline{-12.66}
		\\
		
		\textbf{Full TPDSL}
		&
		\textbf{1413.46}
		&
		\textbf{+40.46}
		&
		\textbf{+0.00}
		\\
		
		\bottomrule
	\end{tabular}
\end{table}

\begin{table}[t]
	\centering
	\small
\caption{
	Effect of different LLM proposers. 
	Qwen-Plus is used as the default proposer in all main experiments, while 
	the other proposers are evaluated only for robustness analysis.
}
	\label{tab:ablation_llm_proposer}
	\setlength{\tabcolsep}{8.0pt}
	\renewcommand{\arraystretch}{1.10}
	
	\begin{tabular}{lc}
		\toprule
		\textbf{LLM Proposer} & \textbf{MME} \\
		\midrule
		Qwen-Max~\cite{bai2023qwen} & 1411.04 \\
		Qwen-Plus~\cite{alibabacloud2026qwenplus} 
		& \underline{1413.46} \\
		DeepSeek-V4-Flash~\cite{xu2026deepseek} 
		& \textbf{1414.34} \\
		\bottomrule
	\end{tabular}
\end{table}

TPDSL decouples the searched policy structure from runtime-dependent fields.
Thus, the selected state $\alpha^\ast$ can be re-materialized for a target 
budget, task, or MLLM backbone as
\begin{equation}
\pi_{\alpha^\ast}^{(K_n)}
=
\mathrm{Materialize}(\alpha^\ast,\pi_{P,n},K_n),
\end{equation}
where only budget- and model-dependent fields are updated. 
This enables transfer without additional LLM-driven search, and each 
transferred policy is validated before deployment.
Algorithm~\ref{alg:tpdsl_residual} summarizes the runtime instantiation and 
execution of the selected TPDSL state.

\section{Experiments and Results}

\subsection{Experimental Protocol}

\paragraph{Benchmarks.}

We evaluate AutoPrune on 14 widely used image-based multimodal benchmarks covering diverse visual understanding capabilities. VQAv2~\cite{goyal2017making}, GQA~\cite{hudson2019gqa}, and VizWiz~\cite{gurari2018vizwiz} assess general visual question answering; ScienceQA-IMG~\cite{lu2022learn} evaluates scientific reasoning; HallBench~\cite{guan2024hallbench} and POPE~\cite{li2023evaluating} measure visual hallucination; and MME~\cite{fu2026mme}, MMBench-EN and MMBench-CN~\cite{liu2025mmbench}, and MM-Vet~\cite{yu2023mmvet} provide comprehensive multimodal evaluation. We further include TextVQA~\cite{singh2019towards}, ChartQA~\cite{masry2022chartqa}, AI2D~\cite{kembhavi2016diagram}, and OCRBench~\cite{liu2024ocrbench} to evaluate text-rich and diagram-oriented visual understanding. All experiments follow the official data splits, evaluation protocols, and metrics of the corresponding benchmarks.

\paragraph{Implementation Details and Evaluation Metrics.}

AutoPrune is training-free and keeps the underlying MLLM frozen throughout 
both policy search and inference. 
LLM-driven policy design is performed once on LLaVA-1.5-7B, using MME as the 
task evaluator and $K_0=32$ as the source visual-token budget. 
The search consists of 10 rounds with five candidate search states per round. 
Unless otherwise specified, we use Qwen-Plus~\cite{alibabacloud2026qwenplus} 
as the LLM proposer and CDPruner~\cite{zhang2025cdpruner} as the base policy 
$\pi_P$. 
The residual exchange parameters are set to $q_e=2$ and $r_{\min}=30$. 
The selected TPDSL search state is subsequently re-instantiated across token 
budgets, evaluation tasks, and MLLM backbones without additional LLM-driven 
search.

We report the official metric for each benchmark, together with the aggregate 
score \emph{Acc.} and the relative performance \emph{Rel.} with respect to the 
corresponding full-token model. 
Among pruning methods, the best and second-best results are highlighted in 
\textbf{bold} and \underline{underlined}, respectively. 
Additional implementation and evaluation details are provided in the 
supplementary material.

\subsection{Main Results}

\paragraph{Comparison with State-of-the-Art Methods.}

We compare AutoPrune with representative training-free visual-token pruning methods on LLaVA-1.5-7B and LLaVA-NeXT-7B. Under the most aggressive setting in Table~\ref{tab:reduction94}, where 94.4\% of visual tokens are removed, AutoPrune achieves aggregate scores of 63.2 and 65.2 on the two backbones, retaining 99.7\% and 99.9\% of their full-token performance, respectively. It outperforms CDPruner by 3.2 points on LLaVA-1.5-7B and 2.4 points on LLaVA-NeXT-7B, with particularly large gains on MME and MMBench.

Figure~\ref{fig:performance-image} compares the methods at token-reduction 
ratios of 77.8\%, 88.9\%, and 94.4\%. 
AutoPrune consistently achieves the strongest aggregate performance, and its 
advantage becomes more evident as the token budget decreases. 
Although the TPDSL state is searched only on MME at the 32-token source budget, 
it remains effective across other benchmarks, budgets, and backbones.
This cross-setting behavior suggests that the searched residual structure 
captures reusable token-selection principles rather than merely overfitting to 
the MME evaluator.
At moderate reduction ratios, AutoPrune even exceeds the corresponding 
full-token models, suggesting that removing redundant visual tokens can reduce 
visual interference while retaining task-relevant information. 
Complete per-benchmark results are provided in the supplementary material.

\paragraph{Efficiency on LLaVA-NeXT-7B.}

Table~\ref{tab:eff} evaluates inference efficiency on LLaVA-NeXT-7B using a single NVIDIA RTX 3090 GPU. With 320 retained visual tokens, AutoPrune reduces FLOPs and prefill latency by $9.9\times$ and $6.4\times$, respectively, while decreasing the KV-cache size from 1440.0\,MB to 160.0\,MB. Compared with CDPruner~\cite{zhang2025cdpruner} under the same token budget, 
AutoPrune preserves the same FLOPs, decoding latency, KV-cache size, and peak 
GPU memory, while achieving comparable prefill latency and improving the MME 
score from 1453.0 to 1457.9. These results show that the searched TPDSL state improves task performance 
without introducing additional inference cost.

\paragraph{Transfer to Advanced MLLM Architectures.}

We further evaluate cross-backbone transfer on 
Qwen2.5-VL-7B~\cite{bai2025qwen25vl}, whose visual-token architecture differs 
substantially from that of the LLaVA models used during search. 
As the official implementation for this setting is unavailable, we reproduce CDPruner~\cite{zhang2025cdpruner} following the evaluation protocol and implementation details reported in its paper.

As shown in Table~\ref{tab:qwen}, AutoPrune consistently improves the 
aggregate score over the reproduced baseline across all evaluated token 
budgets, with particularly strong gains on text-rich visual reasoning tasks.
At 128 tokens, relative performance increases from 74.9\% to 81.0\%, corresponding to a gain of 6.1 percentage points. 
These results demonstrate that the selected TPDSL state can be effectively 
re-instantiated on a distinct visual-token architecture, even under highly 
constrained token budgets.

\subsection{Ablation Studies}
\label{sec:ablation}

Unless specified, all ablations use LLaVA-1.5-7B on MME with 
32 retained visual tokens and a search budget of $10\times5$.

\paragraph{Effect of TPDSL Components.}

We examine three central components of TPDSL: reference anchoring 
(i.e., base-policy preservation), multi-score fusion, and diversity-aware 
selection. 
Budget control and token reassembly are kept fixed because they are necessary 
for executable and budget-compliant pruning. 
The CDPruner base policy achieves an MME score of 1373.00 and serves as the
baseline for the ablation and search-strategy studies.

As shown in Table~\ref{tab:ablation_tpdsl_design}, removing reference 
anchoring causes the largest degradation, reducing MME from 1413.46 to 
1223.29. 
This result confirms that constraining the search state to a residual 
modification is critical for preserving reliable token selections. 
Removing multi-score fusion reduces the score to 1391.70, indicating that a 
single scoring cue is less reliable than combining complementary relevance, 
saliency, spatial, redundancy, and contrast signals. 
Removing diversity-aware selection yields 1400.80, suggesting that controlling 
redundancy among candidate tokens provides an additional benefit. 
The full TPDSL configuration achieves the best result and improves over the 
base policy by 40.46 MME points.

\paragraph{Effect of LLM Proposers.}

We keep Qwen-Plus~\cite{alibabacloud2026qwenplus} as the default proposer for 
all main experiments and evaluate other proposers only for robustness analysis. 
As shown in Table~\ref{tab:ablation_llm_proposer}, the maximum difference among 
the three proposers is only 3.30 MME points. 
Although DeepSeek-V4-Flash~\cite{xu2026deepseek} obtains the highest score, 
Qwen-Plus achieves a closely comparable result of 1413.46. 
This small variation indicates that AutoPrune relies primarily on the 
structured TPDSL space and evaluator feedback rather than on a specific LLM 
proposer. 
Additional multi-seed searches further confirm the search stability of Full 
TPDSL, as detailed in the supplementary material.

\begin{table}[t]
	\centering
	\small
	\caption{
		Effect of $q_e$. $K-q_e$ denotes preserved base-policy tokens;
			$\Delta$ is relative to $q_e=0$.
	}
	\label{tab:ablation_exchange_quota}
	\setlength{\tabcolsep}{4.2pt}
	\renewcommand{\arraystretch}{1.10}
	
	\begin{tabular}{cccc}
		\toprule
			\textbf{$q_e$} & \textbf{$K-q_e$} & \textbf{MME} & \textbf{$\Delta$} \\
		\midrule
0 & 32 & 1373.00 & +0.00 \\
1 & 31 & 1384.95 & +11.95 \\
2 & 30 & \textbf{1413.46} & \textbf{+40.46} \\
4 & 28 & \underline{1394.28} & \underline{+21.28} \\
8 & 24 & 1362.48 & -10.52 \\
16 & 16 & 1326.35 & -46.65 \\
32 & 0  & 1217.88 & -155.12 \\
		\bottomrule
	\end{tabular}
\end{table}

\paragraph{Effect of the Residual Exchange Quota.}

We vary the residual exchange quota $q_e$, which limits the number of 
base-policy tokens replaced by searched candidates. Under the 32-token budget,  the refinement preserves at least $32-q_e$ 
base-policy selections.

As shown in Table~\ref{tab:ablation_exchange_quota}, a small nonzero exchange 
quota improves upon the base-policy-only setting. 
The best result is obtained with $q_e=2$, which improves MME by 40.46 points.
Increasing the quota beyond this value gradually degrades performance, and 
allowing all 32 base-policy tokens to be replaced reduces the score to 1217.88. These results show that residual refinement is most effective when it 
preserves most base-policy selections and corrects only a few uncertain tokens. 
We therefore set $q_e=2$ by default.

\paragraph{Generalization across Token-Pruning Strategies.}
We further instantiate AutoPrune with five alternative token-pruning strategies 
as the base policy $\pi_P$. 
These strategies are implemented under a unified CDPruner-compatible interface 
to isolate their selection or merging priors within the same execution pipeline, 
rather than to exactly reproduce the complete original systems.
This experiment is intended to test whether AutoPrune can refine different 
pruning priors under a controlled interface, rather than to re-rank the 
original systems in their native implementations.

As shown in Table~\ref{tab:ablation_base_policy}, AutoPrune improves all five 
strategy-only baselines by 22.46--145.40 MME points. 
The largest improvements are obtained with PruMerge+ and VisionZip, reaching 
gains of 145.40 and 127.80 points, respectively. 
These consistent gains show that the residual search-state formulation is not 
tied to a particular base policy.

\begin{table}[t]
	\centering
	\scriptsize
	\caption{
		Generalization across base token-pruning strategies on MME at 
		32 tokens. 
		\emph{Strategy only} denotes the corresponding strategy implemented 
		under the unified execution interface, and \emph{Gain} is relative to 
		\emph{Strategy only}.
	}
	\label{tab:ablation_base_policy}
	\setlength{\tabcolsep}{3.4pt}
	\renewcommand{\arraystretch}{1.10}
	
	\begin{tabular}{lccc}
		\toprule
		\textbf{Base Strategy}
		&
		\textbf{Strategy Only}
		&
		\textbf{AutoPrune}
		&
		\textbf{Gain}
		\\
		\midrule
		PruMerge+~\cite{shang2025llava}
		& 1142.82 & 1288.22 & \textbf{+145.40} \\
		
		TRIM~\cite{song2025less}
		& 1514.94 & 1569.54 & +54.60 \\
		
		VisionZip~\cite{yang2025visionzip}
		& 1229.72 & 1357.52 & \underline{+127.80} \\
		
		DART~\cite{wen2025stop}
		& 1488.15 & 1510.61 & +22.46 \\
		
		DivPrune~\cite{alvar2025divprune}
		& \textbf{1574.67} & \textbf{1609.33} & +34.66 \\
		\bottomrule
	\end{tabular}
\end{table}

\section{Conclusion}

We presented \emph{AutoPrune}, a training-free framework that makes 
LLM-driven visual-token pruning design practical through structured residual 
search. 
Rather than asking the LLM to generate unconstrained pruning code, AutoPrune 
uses TPDSL to express candidate policies as executable, budget-compliant, and 
safety-validated modifications to a strong base policy. 
This design preserves reliable prior selections while allowing limited, 
task-adaptive token replacement.

Experiments across multiple benchmarks and MLLM backbones show that AutoPrune 
improves the accuracy--efficiency trade-off over representative handcrafted 
pruning methods, with larger gains under tighter token budgets. 
The results suggest that LLM-driven algorithm design becomes effective for 
visual-token pruning when the search space is expressed as constraint-aware 
residual modifications rather than free-form code.

AutoPrune remains dependent on the expressiveness of the TPDSL search space 
and the reliability of the task evaluator. 
Future work will explore richer search-state representations, more adaptive 
evaluation strategies, and broader validation across MLLM architectures and 
multimodal tasks.

\bibliography{aaai2027}

\appendix
\setcounter{secnumdepth}{1}

\begin{center}
{\Large\bfseries Supplementary Material}\\[0.35em]
{\large\bfseries Towards LLM-Driven Autodesign of Visual Token Pruning Algorithms}
\end{center}

\section{Additional Method Details}
\label{sec:supp_method}

This section provides additional implementation details omitted from the main 
paper. 
We first describe the organization, composition, and runtime semantics of the 
131 atoms in the Token Pruning Domain-Specific Language (TPDSL). 
We then present the complete evaluator-in-the-loop procedure for LLM-driven 
policy design. 
Finally, we report the materialized TPDSL search state selected under the 
source setting.

\subsection{TPDSL Atom Library}
\label{sec:supp_tpdsl_library}

TPDSL represents each candidate pruning design as a structured residual state
\begin{equation}
\mathcal{S}_{\alpha}
=
(\pi_P,\alpha),
\qquad
\alpha=(B,S,C,R),
\end{equation}
where $\pi_P$ is the base policy, and $B$, $S$, $C$, and $R$ denote budget 
control, token scoring, selection constraints, and token reassembly, 
respectively.

The TPDSL library contains 131 instantiated atoms organized into nine 
functional groups:
\begin{equation}
\mathcal{L}
=
\bigcup_{g=1}^{9}
\mathcal{L}^{(g)},
\qquad
|\mathcal{L}|=131,
\label{eq:supp_tpdsl_library}
\end{equation}
where $\mathcal{L}^{(g)}$ denotes the atom set associated with functional 
group $g$.

Table~\ref{tab:tpdsl_library} summarizes the purpose, size, representative 
operations, and search-state role of each group. 
The listed atoms illustrate representative operation families rather than all 
instantiated parameter variants. 
Several selection atoms implement DPP-style or conditional-diversity 
mechanisms motivated by prior work on diverse subset selection and 
visual-token pruning
\cite{chen2018fast,alvar2025divprune,zhang2025cdpruner}. 
Complete machine-readable definitions, parameter ranges, and runtime interfaces 
are provided in the accompanying code package.

\begin{table*}[!t]
	\centering
	\caption{
		Organization of the 131 TPDSL atoms.
		The final column indicates the role of each group in the search state
		$\alpha=(B,S,C,R)$ or in the execution pipeline.
		Implementation identifiers containing \emph{reference} denote 
		base-policy selections.
	}
	\label{tab:tpdsl_library}
	\setlength{\tabcolsep}{4pt}
	\renewcommand{\arraystretch}{1.15}
	\scriptsize

	\begin{tabular}{
			p{2.6cm}
			p{3.0cm}
			c
			p{8.0cm}
			p{1.4cm}
		}
		\toprule
		\textbf{Component Group}
		&
		\textbf{Function}
		&
		\textbf{\# Atoms}
		&
		\textbf{Representative Atoms}
		&
		\textbf{Role}
		\\
		\midrule

		Token Source / Base Policy
		&
		Defines the visual-token source and base-policy context.
		&
		6
		&
		\path{vision_tokens@post_projector},
		\path{vision_tokens@pre_llm},
		\path{external_policy},
		\path{native_teacher_mask_prior},
		\path{reference_mask_prior}.
		&
		Context
		\\

		Budget Control
		&
		Specifies the target token count and budget-instantiation rule.
		&
		10
		&
		Fixed budgets:
		$K\in\{16,32,64,128,160,192,256,320,512,640\}$.
		&
		$B$
		\\

		Scoring Atoms
		&
		Measure token relevance, saliency, representativeness, and redundancy.
		&
		10
		&
		\path{instruction_guided_visual_relevance},
		\path{attention_proxy_saliency},
		\path{spatial_centrality_prior},
		\path{redundancy_density_penalty},
		\path{local_contrast_saliency},
		\path{visual_token_norm_saliency},
		\path{cluster_centroid_representativeness}.
		&
		$S$
		\\

		Similarity / Kernel
		&
		Models pairwise similarity and token redundancy.
		&
		2
		&
		\path{pairwise_cosine_kernel},
		\path{runtime_conditional_similarity_kernel}.
		&
		$S$
		\\

		Score Fusion
		&
		Combines complementary token-quality signals.
		&
		3
		&
		\path{weighted_sum_fusion},
		\path{weighted_product_fusion},
		\path{reference_confidence_weighted_product_fusion}.
		&
		$S$
		\\

		Candidate-Pool Construction
		&
		Constructs and partitions searched candidate-token pools.
		&
		6
		&
		\path{multi_source_pool_builder},
		\path{quality_pool_builder},
		\path{spatial_coverage_pool_builder},
		\path{feature_cluster_partition},
		\path{spatial_grid_partition}.
		&
		$C$
		\\

		Selection Constraints
		&
Enforces reference anchoring, i.e., base-policy preservation, bounded exchange, 
and diversity.
		&
		9
		&
		\path{reference_residual_exchange_selector},
		\path{reference_margin_gated_exchange_selector},
		\path{reference_strict_margin_exchange_selector},
		\path{anchor_controlled_pool_dpp_selector},
		\path{constrained_pool_dpp_selector},
		\path{runtime_reference_anchor_pool_dpp_selector}.
		&
		$C$
		\\

		Reassembly / Token Processing
		&
		Restores retained tokens and optionally incorporates information from 
		discarded tokens.
		&
		81
		&
		\path{nearest_anchor},
		\path{anchor_weighted_residual},
		\path{mean_residual},
		\path{anchor_exchange},
		\path{weighted_average},
		\path{semantic_gated_residual_merge},
		\path{similarity_weighted_residual_merge},
		\path{spatial_local_residual_merge}.
		&
		$R$
		\\

		Safety Checks
		&
		Verifies executability, budget consistency, shape compatibility, and 
		determinism.
		&
		4
		&
		\path{preserve_budget},
		\path{no_shape_change},
		\path{deterministic},
		\path{fallback_to_selection}.
		&
		Validation
		\\

		\midrule
		\textbf{Total} & & \textbf{131} & & \\
		\bottomrule
	\end{tabular}
\end{table*}

\paragraph{Atom Composition and Runtime Semantics.}

A candidate state $\alpha$ selects and parameterizes compatible atoms from the 
TPDSL library. 
The budget component $B$ determines the target token count and resolves 
budget-dependent parameters~\cite{liu2026essence, liu2026structured}. 
The scoring component $S$ composes token-quality signals, similarity kernels, 
and fusion operators. 
The constraint component $C$ constructs the searched candidate pool and 
controls reference anchoring, exchange quotas, and diversity. 
The reassembly component $R$ specifies token ordering, aggregation, and 
shape-preserving output.

The number of atoms in each group reflects the complexity of its operation 
space. In particular, the reassembly and token-processing group contains 81 variants 
covering anchor assignment, residual aggregation, token ordering, and 
shape-preserving output. 
By contrast, similarity kernels, score-fusion operators, and safety checks 
require fewer variants because they implement more standardized operations.

\begin{algorithm}[!t]
	\caption{LLM-Driven TPDSL Policy Design}
	\label{alg:vtp_agent}
	\begin{algorithmic}[1]

		\REQUIRE TPDSL library $\mathcal{L}$, base policy $\pi_P$, 
		source budget $K_0$, task evaluator $\mathcal{E}$, search rounds $T$

		\ENSURE Selected TPDSL state $\alpha^\ast$, executable policy 
		$\pi^\ast$, and evaluation score $E^\ast$

		\STATE Initialize search history
		$\mathcal{H}_0\leftarrow\emptyset$

		\STATE Initialize valid-policy pool
		$\mathcal{V}\leftarrow\emptyset$

		\FOR{$t=1$ to $T$}

			\STATE $\mathcal{A}_t
			\leftarrow
			\mathrm{LLMPropose}
			(\mathcal{L},\pi_P,\mathcal{H}_{t-1};K_0)$

			\STATE Initialize round observations
			$\mathcal{O}_t\leftarrow\emptyset$

			\FOR{each $\alpha_{t,i}\in\mathcal{A}_t$}

				\STATE $\pi_{\alpha_{t,i}}^{(K_0)}
				\leftarrow
				\mathrm{Materialize}
				(\alpha_{t,i},\pi_P,K_0)$

				\STATE $d_{t,i}
				\leftarrow
				\mathrm{SafetyCheck}
				(\pi_{\alpha_{t,i}}^{(K_0)},K_0)$

				\IF{$\mathrm{Valid}(d_{t,i})=0$}

					\STATE $\mathcal{O}_t
					\leftarrow
					\mathcal{O}_t
					\cup
					\{
					(\alpha_{t,i},
					\mathrm{invalid},
					d_{t,i})
					\}$

					\STATE \textbf{continue}

				\ENDIF

				\STATE $e_{t,i}
				\leftarrow
				\mathcal{E}
				(\pi_{\alpha_{t,i}}^{(K_0)})$

				\STATE $\mathcal{O}_t
				\leftarrow
				\mathcal{O}_t
				\cup
				\{
				(\alpha_{t,i},
				e_{t,i},
				d_{t,i})
				\}$

				\STATE $\mathcal{V}
				\leftarrow
				\mathcal{V}
				\cup
				\{
				(\alpha_{t,i},
				\pi_{\alpha_{t,i}}^{(K_0)},
				e_{t,i})
				\}$

			\ENDFOR

			\STATE $\mathcal{H}_t
			\leftarrow
			\mathrm{Summarize}
			(\mathcal{H}_{t-1},\mathcal{O}_t)$

		\ENDFOR

		\STATE $(\alpha^\ast,\pi^\ast,E^\ast)
		\leftarrow
		\underset{(\alpha,\pi,e)\in\mathcal{V}}
		{\arg\max}
		\; e$

		\STATE \textbf{return}
		$(\alpha^\ast,\pi^\ast,E^\ast)$

	\end{algorithmic}
\end{algorithm}

\paragraph{Materialization.}

Given a TPDSL state $\alpha$, base policy $\pi_P$, and token budget $K$, the 
materializer constructs an executable policy:
\begin{equation}
\pi_{\alpha}^{(K)}
=
\mathrm{Materialize}
\left(
\alpha,
\pi_P,
K
\right).
\end{equation}
Materialization binds the selected atoms to their runtime implementations, 
resolves default and budget-dependent parameters, checks inter-component 
compatibility, and constructs the corresponding scoring, selection, exchange, 
and reassembly operators. 
Because the resulting policy is specified through standardized TPDSL 
interfaces, it can be re-instantiated under different token budgets and MLLM 
backbones without modifying the searched policy structure.

\paragraph{Safety Validation.}

Before task evaluation, each materialized policy is subjected to a deterministic 
validation procedure:
\begin{equation}
d_{\alpha}^{(K)}
=
\mathrm{SafetyCheck}
\left(
\pi_{\alpha}^{(K)},
K
\right).
\end{equation}
The diagnostic record $d_{\alpha}^{(K)}$ verifies:

\begin{itemize}
	\item token-budget consistency and exact output cardinality;
	\item validity and uniqueness of token indices;
	\item numerical stability of scoring and fusion operations;
	\item compatibility of tensor shapes and token dimensions; and
	\item deterministic execution under the same input and configuration.
\end{itemize}

A candidate that fails any required check is marked invalid and excluded from 
task evaluation. 
When specified by the TPDSL state, the runtime policy may fall back to the 
base-policy selection rather than returning an invalid token subset.

\subsection{Procedure of LLM-Driven Policy Design}
\label{sec:supp_algo_llm_design}

The main paper summarizes LLM-driven policy design as an iterative 
proposal--validation--evaluation process. 
Algorithm~\ref{alg:vtp_agent} provides the complete procedure. 
At each round, the LLM proposes TPDSL-structured residual states from the 
library, base policy, and accumulated search history. 
Each candidate is materialized and safety-checked before task evaluation. 
Both successful evaluations and validation failures are recorded as feedback 
for subsequent rounds.

The search-history summary $\mathcal{H}_t$ retains the highest-performing 
component combinations, score trends, and recurrent validation failures~\cite{fu2019recognition,fu2023momentum,fu2024understanding,fu2026overcoming}. 
This compressed feedback allows the LLM to exploit effective TPDSL structures 
while avoiding previously observed invalid or ineffective configurations. 
Importantly, the LLM is used only during this offline design stage. 
Once $\alpha^\ast$ has been selected, inference uses the materialized TPDSL 
policy directly and introduces no additional LLM calls.

\subsection{Selected TPDSL State}
\label{sec:supp_selected_tpdsl_state}

Table~\ref{tab:selected_tpdsl_state} summarizes the selected Full TPDSL state 
under the source setting, where LLaVA-1.5-7B is evaluated on MME with 
$K_0=32$ visual tokens. 
The state is expressed as a residual modification to the CDPruner base policy. 
It combines multi-signal token scoring with diversity-aware candidate 
selection and bounded residual exchange, thereby preserving most base-policy 
tokens while revising only a small number of uncertain selections.

The selected state preserves at least 30 of the 32 base-policy tokens and 
permits at most two replacements per input. 
This configuration reflects the central residual-design principle of AutoPrune: 
the base policy supplies the dominant token-selection structure, while the 
searched TPDSL state performs only bounded corrections to uncertain selections.

\begin{table}[!t]
	\centering
	\scriptsize
	\caption{
		Materialized composition of the selected Full TPDSL state under the 
		32-token source setting.
	}
	\label{tab:selected_tpdsl_state}
	\setlength{\tabcolsep}{3.5pt}
	\renewcommand{\arraystretch}{1.10}
	\resizebox{\columnwidth}{!}{
	\begin{tabular}{lll}
		\toprule
		\textbf{Component} &
		\textbf{Materialized Choice} &
		\textbf{Role} \\
		\midrule
		
		$B$ &
		Fixed 32-token budget &
		Budget instantiation \\
		
		\midrule
		\multirow{2}{*}{$S$} &
		Relevance, attention, spatial, redundancy, and contrast signals &
		Multi-score token quality \\
		&
		Weighted product fusion &
		Score aggregation \\
		
		\midrule
		\multirow{3}{*}{$C$} &
		Reference-anchored residual exchange &
		Base-policy preservation \\
		&
		$q_e=2$, $r_{\min}=30$ &
		Bounded token replacement \\
		&
		Diversity-aware candidate selection &
		Redundancy control \\
		
		\midrule
		$R$ &
		Shape-preserving token reassembly &
		Runtime output construction \\
		
		\bottomrule
	\end{tabular}
	}
\end{table}

\section{Experimental Settings}
\label{sec:supp_experiments_settings}

\subsection{MLLM Backbones}
\label{sec:supp_backbones}

We evaluate AutoPrune on three representative MLLM backbones: 
LLaVA-1.5-7B~\cite{liu2024improved}, 
LLaVA-NeXT-7B~\cite{liu2024llavanext}, and 
Qwen2.5-VL-7B~\cite{bai2025qwen25vl}. 
These backbones differ in image-resolution handling, visual-token organization, 
and language-model architecture, enabling evaluation across distinct MLLM 
designs.

For LLaVA-1.5-7B and LLaVA-NeXT-7B, we follow the evaluation protocol 
of~\cite{zhang2025cdpruner}, including the original checkpoints, prompting 
formats, and decoding configurations. 
For Qwen2.5-VL-7B, we preserve its native visual-token architecture and 
reproduce CDPruner following the protocol and implementation details reported 
in~\cite{zhang2025cdpruner}.

\subsection{Search Configuration and Implementation Details}
\label{sec:supp_implementation}

AutoPrune is training-free, with all parameters of the underlying MLLM kept 
frozen. 
Unless otherwise specified, LLM-driven policy design is performed once on 
LLaVA-1.5-7B using MME as the evaluator and a source budget of 
$K_0=32$ visual tokens. 
Qwen-Plus~\cite{bai2023qwen} serves as the default LLM proposer, while 
CDPruner~\cite{zhang2025cdpruner} is used as the base policy $\pi_P$. 
The search runs for 10 rounds with five candidate search states per round, 
yielding 50 candidates in total.

Each candidate search state is materialized into an executable pruning policy 
and validated for token-budget consistency, index validity, numerical 
stability, output-shape compatibility, and deterministic execution. 
Candidates that fail any validation criterion are excluded from task 
evaluation.

For residual refinement, we set the minimum base-policy retention constraint 
to $r_{\min}=30$ and the residual exchange quota to $q_e=2$. 
Under the 32-token source budget, at least 30 base-policy tokens are therefore 
preserved, while at most two selections may be replaced by searched 
candidates.

The selected TPDSL search state is evaluated directly under the source setting 
and re-instantiated for other token budgets and MLLM backbones without 
additional LLM-driven search.

\subsection{Evaluation Metrics and Efficiency Protocol}
\label{sec:supp_metrics}

We report the official metric for each benchmark, together with the aggregate 
score \emph{Acc.} and relative performance \emph{Rel.}. 
Because MME is reported on a substantially larger numerical scale than the 
other benchmarks, it is normalized before aggregation:
\begin{equation}
\widetilde{m}_{\mathrm{MME}}
=
\frac{m_{\mathrm{MME}}}{c_{\mathrm{MME}}},
\label{eq:mme_normalization}
\end{equation}
where $c_{\mathrm{MME}}=20$ for the LLaVA benchmark suite and 
$c_{\mathrm{MME}}=28$ for the Qwen2.5-VL-7B benchmark suite. 
The official scores of all remaining benchmarks are used directly.

The aggregate score is computed as
\begin{equation}
\mathrm{Acc.}
=
\frac{1}{|\mathcal{B}|}
\sum_{b\in\mathcal{B}}
\widetilde{m}_b,
\label{eq:aggregate_accuracy}
\end{equation}
where $\mathcal{B}$ denotes the benchmark set for the corresponding backbone 
and $\widetilde{m}_b$ is the benchmark score after normalization, when 
applicable.

Relative performance is measured against the corresponding full-token model:
\begin{equation}
\mathrm{Rel.}
=
\frac{
	\mathrm{Acc.}_{\mathrm{pruned}}
}{
	\mathrm{Acc.}_{\mathrm{full}}
}
\times 100\%.
\label{eq:relative_accuracy}
\end{equation}
A value above $100\%$ indicates that the pruned model exceeds the aggregate 
performance of its full-token counterpart.

For efficiency evaluation, we report the number of retained visual tokens, 
FLOPs, prefill latency, decoding latency, KV-cache size, and peak GPU memory. 
All methods are evaluated on a single NVIDIA RTX 3090 GPU using identical 
checkpoints, token budgets, and inference configurations. 
Within each token budget, the best and second-best pruning results are marked 
in \textbf{bold} and \underline{underlined}, respectively.

\section{Additional Experimental Results}
\label{sec:supp_complete_results}

This section reports complete per-benchmark results, cross-budget transfer, 
comparisons with alternative LLM-driven search strategies, and search-stability 
analysis across random seeds. 
Unless otherwise specified, AutoPrune uses the TPDSL search state selected on 
LLaVA-1.5-7B at the source budget $K_0=32$. 
For other token budgets and MLLM backbones, only budget- and 
architecture-dependent fields are re-instantiated, without additional 
LLM-driven search.

\subsection{Results on LLaVA-1.5-7B}
\label{sec:supp_llava15_results}

\begin{table*}[!t]
	\centering
	\scriptsize
	\caption{
		Complete results on LLaVA-1.5-7B across visual-token budgets.
		AutoPrune is searched at 32 tokens and re-instantiated at 64 and
		128 tokens without additional LLM-driven search.
		\emph{Acc.} denotes the aggregate score, and \emph{Rel.} is normalized
		to the 576-token full model.
	}
	\label{tab:supp_llava15_full}
	\setlength{\tabcolsep}{3.0pt}
	\renewcommand{\arraystretch}{1.06}
	\resizebox{\linewidth}{!}{
		\begin{tabular}{
				l
				cccccc
				cccc
				cc
			}
			\toprule
\textbf{Method} & \textbf{VQA$^{\mathrm{V2}}$} & \textbf{GQA} &
			\textbf{VizWiz} & \textbf{SQA$^{\mathrm{IMG}}$} &
			\textbf{VQA$^{\mathrm{Text}}$} & \textbf{POPE} &
			\textbf{MME} & \textbf{MMB$^{\mathrm{EN}}$} &
			\textbf{MMB$^{\mathrm{CN}}$} & \textbf{MM-Vet} &
			\textbf{Acc.} & \textbf{Rel.} \\
			\midrule
			\multicolumn{13}{c}{\textit{Full Model: 576 Visual Tokens}} \\
			\midrule
			LLaVA-1.5-7B & 78.50 & 61.90 & 50.10 & 69.50 & 58.20 & 85.90 &
			1506.50 & 64.70 & 58.10 & 31.30 & 63.40 & 100.0 \\
			\midrule
			\multicolumn{13}{c}{
				\textit{Retain 128 Visual Tokens ($\downarrow$77.8\%)}
			} \\
			\midrule
			FastV~(ECCV24) & 71.00 & 54.00 & 51.90 & 69.20 & 56.40 & 68.20 &
			1368.90 & 63.00 & 55.90 & 27.00 & 58.50 & 92.8 \\
			PDrop~(CVPR25) & 74.30 & 57.10 & 49.40 & \textbf{70.10} &
			\underline{56.70} & 77.50 & \textbf{1444.10} & 62.30 & 55.30 &
			27.60 & 60.30 & 95.0 \\
			SparseVLM~(ICML25) & 75.10 & 57.30 & 49.70 & 69.00 & 56.30 &
			83.10 & 1399.30 & 62.60 & 56.90 & 29.70 & 61.00 & 96.3 \\
			PruMerge+~(2024.05) & 75.00 & 58.20 & \underline{53.70} & 69.10 &
			54.00 & 83.10 & 1408.10 & 61.80 & 55.80 & 30.40 & 61.20 & 96.8 \\
			TRIM~(COLING25) & 75.40 & 58.40 & 51.60 & 68.60 & 52.20 & 85.30 &
			1413.40 & 63.00 & 52.30 & 29.90 & 60.70 & 95.8 \\
			VisionZip~(CVPR25) & 75.60 & 57.60 & 51.60 & 68.70 &
			\textbf{56.90} & 83.30 & 1436.90 & 62.10 & 57.00 &
			\underline{31.60} & 61.60 & 97.6 \\
			DART~(2025.02) & 74.70 & 57.90 & 52.80 & 69.10 & 56.30 & 80.40 &
			1408.70 & 60.70 & \underline{57.30} & 30.90 & 61.10 & 96.9 \\
			DivPrune~(CVPR25) & 76.00 & 59.40 & 52.80 & 68.60 & 55.90 &
			87.00 & 1405.10 & 61.50 & 54.80 & 30.60 & 61.70 & 97.5 \\
			CDPruner~(NeurIPS25) & \textbf{76.60} & \textbf{59.90} & 52.80 &
			69.00 & 56.20 & \textbf{87.70} & 1431.40 & \underline{63.10} &
			55.00 & \textbf{32.80} & \underline{62.50} & \underline{99.0} \\
			\textbf{AutoPrune (Ours)} & \textbf{76.60} & \underline{59.82} &
			\textbf{54.25} & \underline{69.84} & 56.27 &
			\underline{87.49} & \underline{1439.28} & \textbf{72.88} &
			\textbf{69.14} & 29.70 & \textbf{64.80} & \textbf{102.3} \\
			\midrule
			\multicolumn{13}{c}{
				\textit{Retain 64 Visual Tokens ($\downarrow$88.9\%)}
			} \\
			\midrule
			FastV~(ECCV24) & 55.90 & 46.00 & 49.10 & \textbf{70.10} & 51.60 &
			35.50 & 973.50 & 50.10 & 42.10 & 18.90 & 46.80 & 74.9 \\
			PDrop~(CVPR25) & 56.30 & 46.10 & 46.30 & 68.80 & 49.20 & 40.80 &
			982.20 & 48.00 & 36.60 & 17.70 & 45.90 & 72.9 \\
			SparseVLM~(ICML25) & 66.90 & 52.00 & 49.40 & 69.20 & 52.10 &
			69.70 & 1190.40 & 58.30 & 49.60 & 24.40 & 55.10 & 87.1 \\
			PruMerge+~(2024.05) & 71.30 & 55.40 & 53.70 & \underline{69.50} &
			52.00 & 75.70 & 1316.80 & 59.60 & 52.10 & 28.00 & 58.30 & 92.4 \\
			TRIM~(COLING25) & 72.40 & 56.60 & 51.10 & 69.00 & 49.70 & 85.90 &
			1350.90 & 60.90 & 48.20 & 24.80 & 58.60 & 91.6 \\
			VisionZip~(CVPR25) & 72.40 & 55.10 & 52.90 & 69.00 &
			\textbf{55.50} & 77.00 & 1365.20 & 60.10 & \underline{55.40} &
			\underline{29.40} & 59.50 & 94.4 \\
			DART~(2025.02) & 71.30 & 54.70 & 53.50 & 69.30 & 54.70 & 73.80 &
			1365.10 & 59.50 & 54.00 & 26.50 & 58.60 & 92.6 \\
			DivPrune~(CVPR25) & 74.10 & 57.50 & \underline{53.60} & 68.00 &
			54.50 & 85.50 & 1334.70 & 60.10 & 52.30 & 28.10 & 60.00 & 94.7 \\
			CDPruner~(NeurIPS25) & \underline{75.40} & \underline{58.60} &
			53.40 & 68.10 & \underline{55.30} & \underline{87.50} &
			\textbf{1415.10} & \underline{61.10} & 53.20 & \textbf{30.50} &
			\underline{61.40} & \underline{97.0} \\
			\textbf{AutoPrune (Ours)} & \textbf{75.50} & \textbf{58.86} &
			\textbf{54.64} & 69.39 & 55.13 & \textbf{87.57} &
			\underline{1406.59} & \textbf{71.19} & \textbf{67.71} & 28.90 &
			\textbf{63.90} & \textbf{100.9} \\
			\midrule
			\multicolumn{13}{c}{
				\textit{Retain 32 Visual Tokens ($\downarrow$94.4\%)}
			} \\
			\midrule
			PruMerge+~(2024.05) & 65.60 & 52.90 & \underline{53.50} & 67.90 &
			49.20 & 66.70 & 1236.60 & 55.10 & 45.90 & 24.70 & 54.30 & 86.1 \\
			TRIM~(COLING25) & 68.60 & 54.50 & 50.70 & 68.10 & 47.60 & 84.90 &
			1251.80 & 57.70 & 40.10 & 20.50 & 55.50 & 86.2 \\
			VisionZip~(CVPR25) & 67.10 & 51.80 & 52.40 & 69.10 & 53.10 &
			69.40 & 1251.20 & 57.00 & \underline{50.30} & 25.30 & 55.80 &
			88.4 \\
			DART~(2025.02) & 67.10 & 52.90 & 52.50 & 69.30 & 52.20 & 69.10 &
			1273.30 & 58.50 & 50.00 & 25.00 & 56.00 & 88.6 \\
			DivPrune~(CVPR25) & 71.20 & 54.90 & 53.30 & 68.60 & 52.90 & 81.50 &
			1284.90 & 57.60 & 49.10 & 26.30 & 58.00 & 91.3 \\
			CDPruner~(NeurIPS25) & \underline{73.60} & \underline{57.00} &
			53.10 & \underline{69.50} & \underline{53.20} & \textbf{87.90} &
			\underline{1373.00} & \underline{59.60} & 49.60 &
			\underline{27.80} & \underline{60.00} & \underline{94.3} \\
			\textbf{AutoPrune (Ours)} & \textbf{73.91} & \textbf{57.33} &
			\textbf{54.27} & \textbf{70.10} & \textbf{53.28} &
			\underline{87.67} & \textbf{1413.46} & \textbf{70.55} &
			\textbf{65.90} & \textbf{28.00} & \textbf{63.20} &
			\textbf{99.7} \\
			\bottomrule
		\end{tabular}
	}
\end{table*}

We compare AutoPrune with representative training-free visual-token pruning 
methods, including FastV~\cite{chen2024image}, 
PyramidDrop (PDrop)~\cite{xing2024pyramiddrop}, 
SparseVLM~\cite{zhang2024sparsevlm}, 
PruMerge+~\cite{shang2025llava}, 
TRIM~\cite{song2025less}, 
VisionZip~\cite{yang2025visionzip}, 
DART~\cite{wen2025stop}, 
DivPrune~\cite{alvar2025divprune}, and 
CDPruner~\cite{zhang2025cdpruner}. 
All methods use the same model checkpoint, benchmark splits, prompting and 
decoding configurations, and visual-token budgets.

Table~\ref{tab:supp_llava15_full} reports complete results at 128, 64, and 
32 retained visual tokens. 
AutoPrune selects a TPDSL search state only at the 32-token source budget and 
re-instantiates it at 64 and 128 tokens without additional search. 
Compared with CDPruner, AutoPrune improves the aggregate score by 2.3, 2.5, 
and 3.2 points at 128, 64, and 32 tokens, respectively. 
The corresponding relative-performance gains are 3.3, 3.9, and 5.4 
percentage points.

The advantage becomes larger under tighter token budgets. 
At 32 tokens, AutoPrune achieves an aggregate score of 63.2 and preserves 
99.7\% of full-token performance. 
The consistent gains at transferred budgets further show that the selected 
search state generalizes beyond its source budget.

\subsection{Results on LLaVA-NeXT-7B}
\label{sec:supp_llavanext_results}

\begin{table*}[!t]
	\centering
	\scriptsize
	\caption{
		Complete results on LLaVA-NeXT-7B across visual-token budgets.
		The TPDSL search state selected on LLaVA-1.5-7B is transferred and
		re-instantiated at 160, 320, and 640 tokens without additional
		LLM-driven search. \emph{Acc.} denotes the aggregate score, and
		\emph{Rel.} is normalized to the 2,880-token full model.
	}
	\label{tab:supp_llavanext_full}
	\setlength{\tabcolsep}{3.0pt}
	\renewcommand{\arraystretch}{1.06}
	\resizebox{\linewidth}{!}{
		\begin{tabular}{
				l
				cccccc
				cccc
				cc
			}
			\toprule
\textbf{Method} & \textbf{VQA$^{\mathrm{V2}}$} & \textbf{GQA} &
			\textbf{VizWiz} & \textbf{SQA$^{\mathrm{IMG}}$} &
			\textbf{VQA$^{\mathrm{Text}}$} & \textbf{POPE} &
			\textbf{MME} & \textbf{MMB$^{\mathrm{EN}}$} &
			\textbf{MMB$^{\mathrm{CN}}$} & \textbf{MM-Vet} &
			\textbf{Acc.} & \textbf{Rel.} \\
			\midrule
			\multicolumn{13}{c}{\textit{Full Model: 2,880 Visual Tokens}} \\
			\midrule
			LLaVA-NeXT-7B & 81.3 & 62.5 & 55.2 & 67.5 & 60.3 & 86.8 &
			1511.8 & 65.8 & 57.3 & 40.0 & 65.2 & 100.0 \\
			\midrule
			\multicolumn{13}{c}{
				\textit{Retain 640 Visual Tokens ($\downarrow$77.8\%)}
			} \\
			\midrule
			FastV~(ECCV24) & 77.0 & 58.9 & 53.9 & 67.4 & 58.1 & 79.5 &
			1412.6 & 63.1 & 53.5 & \underline{39.5} & 62.2 & 95.6 \\
			PDrop~(CVPR25) & 79.1 & 60.0 & 53.8 & 66.7 & 57.8 & 83.8 &
			1475.9 & 64.1 & 55.2 & 36.7 & 63.1 & 96.5 \\
			SparseVLM~(ICML25) & 79.2 & 61.2 & 53.6 & 67.6 &
			\underline{59.7} & 85.3 & 1456.8 & 65.9 & \underline{58.6} &
			36.1 & 64.0 & 97.9 \\
			PruMerge+~(2024.05) & 78.2 & 60.8 & \textbf{57.9} & 67.8 & 54.9 &
			85.3 & \underline{1480.2} & 64.6 & 57.3 & 32.7 & 63.4 & 96.6 \\
			TRIM~(COLING25) & 78.3 & \underline{62.1} & 54.8 & 66.9 & 54.8 &
			86.9 & 1471.8 & \underline{66.8} & 55.8 & 37.8 & 63.8 & 97.6 \\
			VisionZip~(CVPR25) & 79.1 & 61.2 & \underline{57.1} &
			\underline{68.1} & \textbf{59.9} & 86.0 & \textbf{1493.4} &
			65.8 & 58.1 & 38.9 & 64.9 & 99.5 \\
			DART~(2025.02) & 78.3 & 61.3 & 57.0 & \textbf{68.2} & 59.5 &
			85.0 & 1450.2 & 64.9 & 57.1 & 36.9 & 64.1 & 98.2 \\
			DivPrune~(CVPR25) & \underline{79.3} & 61.9 & 55.7 & 67.8 & 57.0 &
			86.9 & 1469.7 & 65.8 & 57.3 & 38.0 & 64.3 & 98.5 \\
			CDPruner~(NeurIPS25) & \textbf{79.9} & \textbf{62.6} & 55.6 &
			67.9 & 58.4 & \textbf{87.3} & 1474.5 & 66.3 & 57.5 &
			\textbf{41.9} & \underline{65.1} & \underline{100.1} \\
			\textbf{AutoPrune (Ours)} & \textbf{79.9} & \textbf{62.6} & 56.9 &
			68.0 & 58.4 & \underline{87.1} & 1466.2 & \textbf{74.9} &
			\textbf{71.1} & 36.1 & \textbf{66.8} & \textbf{102.5} \\
			\midrule
			\multicolumn{13}{c}{
				\textit{Retain 320 Visual Tokens ($\downarrow$88.9\%)}
			} \\
			\midrule
			FastV~(ECCV24) & 61.5 & 49.8 & 51.3 & 66.6 & 52.2 & 49.5 &
			1099.0 & 53.4 & 42.5 & 20.0 & 50.2 & 76.9 \\
			PDrop~(CVPR25) & 66.8 & 50.4 & 49.7 & 66.7 & 49.0 & 60.8 &
			1171.5 & 55.5 & 44.7 & 24.0 & 52.6 & 80.3 \\
			SparseVLM~(ICML25) & 74.6 & 57.9 & 54.2 & 67.2 & 56.5 & 76.9 &
			1386.1 & 63.1 & 56.7 & 32.8 & 60.9 & 93.3 \\
			PruMerge+~(2024.05) & 75.3 & 58.8 & \textbf{57.7} &
			\textbf{68.1} & 54.0 & 79.5 & 1444.3 & 63.0 & 55.6 & 31.4 &
			61.6 & 94.0 \\
			TRIM~(COLING25) & 74.9 & 59.9 & 53.5 & 66.2 & 50.2 & 86.5 &
			1443.8 & 63.5 & 51.0 & 32.7 & 61.1 & 92.9 \\
			VisionZip~(CVPR25) & 76.2 & 58.9 & 56.2 & 67.5 &
			\textbf{58.8} & 82.3 & 1397.1 & 63.3 & 55.6 & 35.8 & 62.4 &
			95.7 \\
			DART~(2025.02) & 75.7 & 59.5 & \underline{56.8} & 67.5 &
			\underline{57.6} & 81.0 & 1419.5 & 64.2 & \underline{55.7} &
			35.7 & 62.5 & 95.8 \\
			DivPrune~(CVPR25) & 77.2 & \underline{61.1} & 55.6 & 67.7 & 56.2 &
			84.7 & 1423.3 & 63.9 & \underline{55.7} & 34.8 & 62.8 & 96.0 \\
			CDPruner~(NeurIPS25) & \underline{78.4} & \textbf{61.6} & 55.8 &
			\underline{67.8} & 57.4 & \textbf{87.2} & \underline{1453.0} &
			\underline{65.5} & \underline{55.7} & \textbf{37.9} &
			\underline{64.0} & \underline{98.0} \\
			\textbf{AutoPrune (Ours)} & \textbf{78.6} & \textbf{61.6} & 56.6 &
			67.7 & 55.6 & \underline{87.1} & \textbf{1457.9} &
			\textbf{73.9} & \textbf{69.1} & \underline{36.0} &
			\textbf{65.9} & \textbf{101.0} \\
			\midrule
			\multicolumn{13}{c}{
				\textit{Retain 160 Visual Tokens ($\downarrow$94.4\%)}
			} \\
			\midrule
			PruMerge+~(2024.05) & 70.5 & 56.2 & \textbf{57.2} & 66.9 & 50.3 &
			71.1 & 1289.6 & 58.0 & 48.9 & 29.3 & 57.3 & 87.7 \\
			TRIM~(COLING25) & 71.0 & 57.4 & 52.9 & 65.5 & 45.8 &
			\underline{84.8} & 1275.8 & 61.6 & 45.2 & 29.6 & 57.8 & 87.7 \\
			VisionZip~(CVPR25) & 71.4 & 55.2 & 55.5 & \textbf{67.9} & 55.0 &
			74.9 & 1327.8 & 58.6 & 50.4 & 32.3 & 58.8 & 90.0 \\
			DART~(2025.02) & 72.5 & 56.8 & \underline{56.7} &
			\underline{67.8} & 54.9 & 75.3 & 1325.4 & 62.0 & 53.6 & 32.2 &
			59.8 & 91.7 \\
			DivPrune~(CVPR25) & 75.0 & 59.3 & 56.1 & 67.1 & 54.1 & 80.0 &
			1356.6 & 62.9 & 53.7 & 32.0 & 60.8 & 92.9 \\
			CDPruner~(NeurIPS25) & \underline{76.7} & \textbf{60.8} & 55.2 &
			67.5 & \underline{55.4} & \textbf{86.8} & \underline{1425.3} &
			\underline{64.2} & \underline{53.8} & \textbf{36.2} &
			\underline{62.8} & \underline{96.0} \\
			\textbf{AutoPrune (Ours)} & \textbf{76.9} & \underline{60.7} &
			56.3 & 67.4 & \textbf{55.6} & \textbf{86.8} &
			\textbf{1430.4} & \textbf{73.3} & \textbf{67.9} &
			\underline{35.3} & \textbf{65.2} & \textbf{99.9} \\
			\bottomrule
		\end{tabular}
	}
\end{table*}

We evaluate cross-backbone transfer on LLaVA-NeXT-7B using 640, 320, and 
160 retained visual tokens. 
The TPDSL search state selected on LLaVA-1.5-7B at $K_0=32$ is transferred 
and re-instantiated for each target budget without additional LLM-driven 
search.

As shown in Table~\ref{tab:supp_llavanext_full}, AutoPrune improves the 
aggregate score over CDPruner by 1.7, 1.9, and 2.4 points at 640, 320, and 
160 tokens, respectively. 
The corresponding relative-performance gains are 2.4, 3.0, and 3.9 
percentage points. 
At the most aggressive 160-token setting, AutoPrune preserves 99.9\% of 
full-token performance.

These consistent gains show that the selected TPDSL search state is not tied 
to the source backbone or its visual-token organization. 
The search-state structure can be transferred while re-instantiating only 
budget- and architecture-dependent fields.

\subsection{Cross-Budget Policy Transfer}
\label{sec:supp_budget_transfer}

\begin{table*}[!t]
	\centering
	\scriptsize
	\caption{
		Cross-budget transfer on LLaVA-1.5-7B. 
		The TPDSL search state selected at 32 tokens is re-instantiated at 
		16, 64, and 128 tokens without additional LLM-driven search.
	}
	\label{tab:transfer}
	\setlength{\tabcolsep}{4.0pt}
	\renewcommand{\arraystretch}{1.12}
	\resizebox{\linewidth}{!}{
		\begin{tabular}{
				c
				cccccc
				cccc
				cc
			}
			\toprule
			\textbf{\# Tokens} & \textbf{VQA$^{\mathrm{V2}}$} & \textbf{GQA} &
			\textbf{VizWiz} & \textbf{SQA$^{\mathrm{IMG}}$} &
			\textbf{VQA$^{\mathrm{Text}}$} & \textbf{POPE} &
			\textbf{MME} & \textbf{MMB$^{\mathrm{EN}}$} &
			\textbf{MMB$^{\mathrm{CN}}$} & \textbf{MM-Vet} &
			\textbf{Acc.} & \textbf{Rel.} \\
			\midrule
			\multicolumn{13}{c}{\textit{Full Model: 576 Visual Tokens}} \\
			\midrule
			576 & 78.50 & 61.90 & 50.10 & 69.50 & 58.20 & 85.90 &
			1506.50 & 64.70 & 58.10 & 31.30 & 63.40 & 100.0 \\
			\midrule
			\multicolumn{13}{c}{\textit{Cross-Budget Transfer of AutoPrune}} \\
			\midrule
			16 & 71.14 & 55.33 & 52.90 & 69.79 & 51.00 & 86.90 &
			1314.29 & 69.84 & 61.82 & 27.80 & 61.20 & 96.6 \\
			32 & 73.91 & 57.33 & 54.27 & 70.10 & 53.28 & 87.67 &
			1413.46 & 70.55 & 65.90 & 28.00 & 63.20 & 99.7 \\
			64 & 75.50 & 58.86 & 54.64 & 69.39 & 55.13 & 87.57 &
			1406.59 & 71.19 & 67.71 & 28.90 & 63.90 & 100.9 \\
			128 & 76.60 & 59.82 & 54.25 & 69.84 & 56.27 & 87.49 &
			1439.28 & 72.88 & 69.14 & 29.70 & 64.80 & 102.3 \\
			\bottomrule
		\end{tabular}
	}
\end{table*}

We evaluate whether the TPDSL search state selected at the 32-token source 
budget generalizes to other token budgets. 
As shown in Table~\ref{tab:transfer}, the 32-token result uses the selected 
source policy directly, while the 16-, 64-, and 128-token results are obtained 
by re-instantiating its budget-dependent fields without additional search.

The selected search state remains effective across all evaluated budgets. 
With only 16 retained tokens, AutoPrune preserves 96.6\% of full-token 
performance. 
As the budget increases from 16 to 128 tokens, the aggregate score rises from 
61.2 to 64.8, and relative performance increases from 96.6\% to 102.3\%. 
These results demonstrate effective cross-budget transfer without repeated 
policy design.

\subsection{Comparison with Different Search Strategies}
\label{sec:supp_search_strategy}

\begin{table}[!t]
	\centering
	\small
	\caption{
		Comparison with different search strategies on 
		LLaVA-1.5-7B using MME. 
		All methods retain 32 visual tokens and use 10 rounds with five 
		candidates per round. 
		$\Delta$ is measured relative to the CDPruner base policy.
	}
	\label{tab:search_strategy_comparison}
	\setlength{\tabcolsep}{5.0pt}
	\renewcommand{\arraystretch}{1.10}
	\begin{tabular}{lccc}
		\toprule
		\textbf{Search Strategy} &
		\textbf{Search Budget} &
		\textbf{MME} &
		\textbf{$\Delta$} \\
		\midrule
		CDPruner base policy & -- & 1373.00 & +0.00 \\
		OpenEvolve & $10 \times 5$ & 1407.23 & +34.23 \\
		EvoPrompting & $10 \times 5$ &
		\underline{1409.21} & \underline{+36.21} \\
		\textbf{AutoPrune (Ours)} & $10 \times 5$ &
		\textbf{1413.46} & \textbf{+40.46} \\
		\bottomrule
	\end{tabular}
\end{table}

We compare AutoPrune with OpenEvolve, an AlphaEvolve-style coding-agent 
baseline~\cite{novikov2025alphaevolve}, and an EvoPrompting-style 
implementation~\cite{chen2023evoprompting}. 
For a controlled comparison, all methods start from the same 
CDPruner~\cite{zhang2025cdpruner} base policy and use the same evaluator, 
32-token budget, and search budget of 10 rounds with five candidates per 
round.

As shown in Table~\ref{tab:search_strategy_comparison}, all search strategies 
improve upon the CDPruner base policy. 
AutoPrune achieves the best MME score of 1413.46, corresponding to a 
40.46-point improvement. 
It outperforms OpenEvolve and EvoPrompting-style search by 6.23 and 4.25 
points, respectively.

Under the same experimental configuration, AutoPrune achieves the best result 
among the evaluated search strategies, supporting the effectiveness of its 
TPDSL-based residual search-state formulation for visual-token pruning.

\subsection{Search Stability across Random Seeds}

\begin{table}[!t]
	\centering
	\scriptsize
	\caption{
		Search stability of Full TPDSL across different random seeds on 
		LLaVA-1.5-7B using MME. 
		All runs retain 32 visual tokens and use a 10$\times$5 search budget. Valid denotes the number of executable and budget-compliant candidates.
Mean, Min, and Std are computed over the 50 candidates within each search run.
	}
	\label{tab:full_tpdsl_multiseed}
	\setlength{\tabcolsep}{3.5pt}
	\renewcommand{\arraystretch}{0.9}
	\resizebox{\columnwidth}{!}{
	\begin{tabular}{cccccc}
		\toprule
		\textbf{Seed} &
		\textbf{Valid} &
		\textbf{Best} &
		\textbf{Mean} &
		\textbf{Min} &
		\textbf{Std} \\
		\midrule
		43 & 50 & 1413.46 & 1408.23 & 1406.82 & 0.89 \\
		44 & 50 & 1413.46 & 1408.62 & 1406.82 & 1.69 \\
		45 & 50 & 1413.46 & 1408.94 & 1407.57 & 1.88 \\
		46 & 50 & 1413.46 & 1408.67 & 1406.82 & 1.86 \\
		\bottomrule
	\end{tabular}
	}
\end{table}

To further examine the stability of Full TPDSL search, we repeat the search 
with four different random seeds under the same experimental setting. 
Each run uses a 10$\times$5 search budget and evaluates 50 candidate policies. 
As shown in Table~\ref{tab:full_tpdsl_multiseed}, all four runs produce 
50 valid candidates, indicating that the TPDSL search space can consistently 
generate executable and budget-compliant pruning policies.

In terms of performance, all four seeds recover the same best MME score of 1413.46. 
The candidate-level mean scores are also close across seeds, ranging from 
1408.23 to 1408.94, and the minimum candidate score remains above 1406.82. 
These results suggest that Full TPDSL does not rely on a lucky high-performing 
proposal; instead, the constrained residual-search space consistently yields 
high-quality candidate policies.

Together with the component ablation in the main paper, these results indicate 
that the gain is not due to an isolated lucky proposal, but is supported by the 
structured TPDSL representation and its constrained residual-search design.

\end{document}